%% file: rmdasparse.tex
\def\arxiv{1}
\documentclass{article} 
\usepackage{iclr2022_conference,times}

\input{math_commands.tex}

\usepackage[toc,page,header]{appendix}
\usepackage{minitoc}

\usepackage{natbib}
\usepackage{paralist}
\usepackage{enumerate}
\usepackage[pagebackref=true]{hyperref}
\usepackage{enumitem}
\usepackage[ruled,noend]{algorithm2e}
\usepackage{url}
\usepackage{subcaption}
\usepackage{graphicx}
\graphicspath{}
\ifdefined\arxiv
\usepackage[margin=0.8in]{geometry}
\fi
\usepackage{amsfonts,xspace}
\usepackage{lmodern}
\usepackage{siunitx}
\usepackage{booktabs}
\usepackage{etoolbox}
\SetKwInOut{Input}{input}
\SetKwInOut{Output}{output}
\SetKwComment{Comment}{// }{}

\usepackage[capitalize,nameinlink]{cleveref}
\crefformat{equation}{\textup{#2(#1)#3}}
\crefrangeformat{equation}{\textup{#3(#1)#4--#5(#2)#6}}
\crefmultiformat{equation}{\textup{#2(#1)#3}}{ and \textup{#2(#1)#3}}
{, \textup{#2(#1)#3}}{, and \textup{#2(#1)#3}}
\crefrangemultiformat{equation}{\textup{#3(#1)#4--#5(#2)#6}}%
{ and \textup{#3(#1)#4--#5(#2)#6}}{, \textup{#3(#1)#4--#5(#2)#6}}{, and \textup{#3(#1)#4--#5(#2)#6}}

\Crefformat{equation}{#2Equation~\textup{(#1)}#3}
\Crefrangeformat{equation}{Equations~\textup{#3(#1)#4--#5(#2)#6}}
\Crefmultiformat{equation}{Equations~\textup{#2(#1)#3}}{ and \textup{#2(#1)#3}}
{, \textup{#2(#1)#3}}{, and \textup{#2(#1)#3}}
\Crefrangemultiformat{equation}{Equations~\textup{#3(#1)#4--#5(#2)#6}}%
{ and \textup{#3(#1)#4--#5(#2)#6}}{, \textup{#3(#1)#4--#5(#2)#6}}{, and \textup{#3(#1)#4--#5(#2)#6}}

\crefdefaultlabelformat{#2\textup{#1}#3}
\usepackage{amsmath,amsthm,mathtools,amssymb}
\newtheorem{assumption}{Assumption}
\newtheorem{definition}{Definition}
\newtheorem{lemma}{Lemma}

\newtheorem{theorem}{Theorem}
\crefname{assumption}{Assumption}{Assumptions}

\DeclareMathOperator{\MCP}{\text{MCP}}
\newcommand*{\lmss}{\fontfamily{lmss}\selectfont}
\newcommand{\norm}[1]{{\left\|{#1}\right\|}}
\newcommand{\prox}{\mbox{\rm prox}}
\def\rmda{{\lmss RMDA}\xspace}
\def\rigl{{\lmss RigL}\xspace}
\def\sgd{{\lmss ProxSGD}\xspace}
\def\ssi{{\lmss ProxSSI}\xspace}
\newcommand{\inprod}[2]{\langle#1,\,#2\rangle}
\newcommand{\as}{\xrightarrow{\text{a.s.}}}

\title{Training Structured Neural Networks Through Manifold
Identification and Variance Reduction}

\iclrfinalcopy

\author{Zih-Syuan Huang\\
Academia Sinica\\
\texttt{zihsyuan@stat.sinica.edu.tw}
\And
Ching-pei Lee\\
Academia Sinica\\
\texttt{leechingpei@gmail.com}
}

\begin{document}
\doparttoc 
\faketableofcontents 

\part{} 

\maketitle

\begin{abstract}
This paper proposes an algorithm, \rmda, for training neural networks
(NNs) with a regularization term for promoting desired structures.
\rmda does not incur computation additional to proximal SGD with
momentum, and achieves variance reduction without requiring the
objective function to be of the finite-sum form.
Through the tool of manifold identification from nonlinear
optimization, we prove that after a finite number of
iterations, all iterates of \rmda possess a desired structure
identical to that induced by the regularizer at the stationary point
of asymptotic convergence, even in the presence of engineering tricks
like data augmentation that complicate the training process.
Experiments on training NNs with structured sparsity confirm that
variance reduction is necessary for such an identification, and
show that \rmda thus significantly outperforms existing methods
for this task.
For unstructured sparsity, \rmda also outperforms a state-of-the-art
pruning method, validating the benefits of training structured NNs
through regularization.
Implementation of \rmda is available at
\url{https://www.github.com/zihsyuan1214/rmda}.
\end{abstract}

\section{Introduction}
\label{sec:intro}
Training neural networks (NNs) with regularization to obtain a certain
desired structure such as structured sparsity or discrete-valued
parameters is a problem of increasing interest.
Existing approaches either use stochastic subgradients of the
regularized objective \citep{wen2016learning,WW18a} or combine popular
stochastic gradient algorithms for NNs, like SGD with momentum (MSGD)
or Adam \citep{DPK15a}, with the proximal operator associated with the
regularizer to conduct proximal stochastic gradient updates to obtain
a model with preferred structures
\citep{bai2018proxquant,yang2019proxsgd,JY20a,deleu2021structured}.
Such methods come with proven convergence for certain measures of
first-order optimality and have shown some empirical success in
applications.
However, we notice that an essential theoretical support lacking in
existing methods is the guarantee for the output iterate to possess
the same structure as that at the point of convergence.
More specifically, often the imposed regularization is only known to
induce a desired structure exactly at optimal or stationary points
of the underlying optimization problem \citep[see for
example,][]{PZ06a}, but training algorithms are only able to generate
iterates asymptotically converging to a stationary point.
Without further theoretical guarantees, it is unknown whether the
output iterate, which is just an approximation of the stationary
point, still has the same structure.
For example, let us assume that sparsity is desired, the point of
convergence is $x^* = (1,0,0)$,
and two algorithms respectively produce iterates $\{y^t =
	(1,t^{-1},t^{-1})\}$ and $\{z^t = (1+t^{-1},0,0)\}$.
Clearly, both iterate sequences converge to $x^*$, but only $z^t$ has
the same desired structure as its limit point $x^*$, while $y^t$ is not
useful for sparsity despite that the point of convergence is.
This work aims at filling this gap to propose an algorithm for
training structured NNs that can provably make all its iterates after
a finite number of iterations possess the desired structure of the
stationary point to which the iterates converge.
We term the structure at a stationary point a stationary structure,
and it should be understood that for multiple stationary points, each
might correspond to a different stationary structure, and we aim at
identifying the one at the limit point of the iterates of an
algorithm, instead of selecting the optimal one among all stationary
structures.
Although finding the structure at an inferior stationary point might
seem not very meaningful,
another reason for studying this identification property is that for
the same point of convergence, the structure at the limit point is the
most preferable one.
Consider the same example above, we note that for any sequence $\{x^t\}$
converging to $x^*$, $x^t_1 \neq 0$ for all $t$ large enough, for
otherwise $x^t$ does not converge to $x^*$.
Therefore, $x^t$ cannot be sparser than $x^*$ if $x^t \rightarrow
x^*$.\footnote{See a more detailed discussion in \cref{app:sparse}.}
Identifying the structure of the point of convergence thus also
amounts to finding the locally most ideal structure under the same
convergence premise.

It is well-known in the literature of nonlinear optimization that
generating iterates consistently possessing the structure at the
stationary point of convergence is possible if all points with the
same structure near the stationary point can be presented locally as
a manifold along which the regularizer is smooth.
This manifold is often termed as the active manifold (relative to the
given stationary point), and the task of generating iterates staying in
the active manifold relative to the point of convergence after finite
iterations is called manifold identification
\citep{Lew02a,HarL04a,LewZ13a}.
To identify the active manifold of a stationary point, we need the
regularizer to be partly smooth \citep{Lew02a,HarL04a} at that point,
roughly meaning that the regularizer is smooth along the active
manifold around the point, while the change in its value is drastic
along directions leaving the manifold.
A more technical definition will be given in \cref{sec:analysis}.
Fortunately, most regularizers used in machine learning are
partly smooth, so stationary structure identification is possible, and
various deterministic algorithms are known to achieve so
\citep{HarL07a,Har11a,Wri12a,LiaFP17a,LiaFP17b,YSL20a,CPL20b,GB20a}.

On the other hand, for stochastic gradient-related methods to identify
a stationary structure, existing theory suggests that the variance of
the gradient estimation needs to vanish as the iterates approach a
stationary point \citep{PooLS18a},
and indeed, it is observed empirically that proximal stochastic
gradient descent (SGD) is incapable of manifold identification due to
the presence of the variance in the gradient estimation
\citep{LeeW12a,SunJNS19a}.\footnote{An exception is the interpolation
case, in which the variance of plain SGD vanishes asymptotically.  But
data augmentation often fails this interpolation condition.}
\citet{PooLS18a} showed that variance-reduction methods such as SVRG
\citep{RJ13a,LX14a} and SAGA \citep{AD14a} that utilize the finite-sum
structure of empirical risk minimization to drive the variance
of their gradient estimators to zero are suitable for this task.
Unfortunately, with the standard practice of data augmentation
in deep learning, training of deep learning models with a
regularizer should be treated as the following stochastic
optimization problem that minimizes the expected loss over a
distribution, instead of the commonly seen finite-sum form:
\begin{equation}
\min_{W \in \gE}\quad F\left( W \right) \coloneqq \E_{\xi\sim \gD}
\left[ f_{\xi} \left( W \right) \right] + \psi\left( W \right),
\label{eq:f}
\end{equation}
where $\gE$ is a Euclidean space with inner product
$\inprod{\cdot}{\cdot}$ and the associated norm $\norm{\cdot}$,  $\gD$
is a distribution over a space $\Omega$, $f_\xi$ is differentiable
almost everywhere for all $\xi \in \Omega$, and $\psi(W)$ is a
regularizer that might be nondifferentiable.
We will also use the notation $f(W) \coloneqq \E_{\xi \sim
\gD}[f_\xi(W)]$.
Without a finite-sum structure in \cref{eq:f}, \citet{AD19a} pointed
out that classical variance-reduction methods are ineffective for deep
learning, and one major reason is the necessity of periodically
evaluating $\nabla f(W)$ (or at least using a large batch from $\gD$ to get a
precise approximation of it) in variance-reduction methods is
intractable, hence manifold identification and therefore finding
the stationary structure becomes an extremely tough task for deep
learning.
Although recently there are efforts in developing variance-reduction
methods for \cref{eq:f} inspired by online problems
\citep{ZW19a,LMN21a,NHP20a,AC19a,QTD19a}, these methods
all have multiple hyperparameters to tune and incur computational cost
at least twice or thrice to that of (proximal) SGD.
As the training of deep learning models is time- and resource-consuming,
these drawbacks make such methods less ideal for deep learning.

To tackle these difficulties, we extend the recently proposed
modernized dual averaging framework \citep{JS20a} to the regularized
setting by incorporating proximal operations, and obtain a new
algorithm \rmda (Regularized Modernized Dual Averaging) for
\cref{eq:f}.
The proposed algorithm provably achieves variance reduction beyond
finite-sum problems without any cost or hard-to-tune hyperparameters
additional to those of proximal momentum SGD (proxMSGD),
and we provide theoretical guarantees for its convergence and
ability for manifold identification.
The key difference between \rmda and the original regularized dual
averaging (RDA) of \citet{LX10a} is that \rmda
incorporates momentum and can achieve better performance for deep
learning in terms of the generalization ability, 
and the new algorithm requires nontrivial proofs for its guarantees.
We further conduct experiments on training deep learning models with a
regularizer for structured-sparsity to demonstrate the ability of
\rmda to identify the stationary structure without sacrificing the
prediction accuracy.

When the desired structure is (unstructured) sparsity, a popular
approach is pruning that trims a given dense model to a specified
level, and works like \citep{TG19a,DB20a,UE20a,SV21a} have shown
promising results.
However, as a post-processing approach, pruning is essentially
different from structured training considered in this work,
because pruning is mainly used when a model is available,
while structured training combines training and structure inducing in
one procedure to potentially reduce the computational cost and memory
footprint when resources are scarce.
We will also show in our experiment that \rmda can achieve better
performance than a state-of-the-art pruning method, suggesting that
structured training indeed has its merits for obtaining sparse NNs.

The main contributions of this work are summarized as follows.
\vspace{-7pt}
\begin{itemize}[leftmargin=*]
\itemsep 2pt
\topskip 0pt
	\item \textbf{\emph{Principled analysis}}: We use the theory of
		manifold identification from
		nonlinear optimization to provide a unified way towards better
		understanding of algorithms for training structured neural
		networks.
	\item \textbf{\emph{Variance reduction beyond finite-sum with low
		cost}}: \rmda achieves variance reduction for problems that
		consist of an infinite-sum term plus a regularizer (see
		\cref{lemma:iterate}) while incorporating momentum to improve
		the generalization performance.
		Its spatial and computational cost is almost the same as
		proxMSGD, and there is no additional hyperparameters to tune,
		making \rmda suitable for large-scale deep learning.
	\item \textbf{\emph{Structure identification}}: With
		the help of variance reduction, our theory shows that under
		suitable conditions, after a finite number of iterations,
		iterates of \rmda stay in the active manifold of its limit point.
	\item \textbf{\emph{Superior empirical performance}}: Experiments
		on neural networks with structured sparsity exemplify that \rmda
		can identify a stationary structure without reducing the
		validation accuracy, thus outperforming existing methods by
		achieving higher group sparsity.
		Another experiment on unstructured sparsity also shows \rmda
		outperforms a state-of-the-art pruning method.

\end{itemize}
\vspace{-7pt}

After this work is finished, we found a very recent paper \cite{VK21a}
that proposed the same algorithm (with slightly differences in the
parameters setting in \cref{eq:prox} of \cref{alg:rmda}) and analyzed
the expected convergence of \cref{eq:f} under a specific scheduling
of $c_t = s_{t+1} \alpha_{t+1}^{-1}$ when both terms are convex.
In contrast, our work focuses on nonconvex deep learning problems, and
especially on the manifold identification aspect.

\section{Algorithm}
Details of the proposed \rmda are in \cref{alg:rmda}.
At the $t$-th iteration with the iterate $W^{t-1}$,  we draw an
independent sample $\xi_t \sim \gD$ to compute the stochastic gradient
$\nabla f_{\xi_t}(W^{t-1})$, decide a learning rate $\eta_t$,
and update the weighted sum $V_t$ of previous stochastic
gradients using $\eta_t$ and the scaling factor $\beta_t \coloneqq
\sqrt{t}$:
\[
	V_0 \coloneqq 0,\quad V_t \coloneqq \sum\nolimits_{k=1}^t \eta_k
	\beta_k \nabla f_{\xi_k}(W^{k-1}) = V_{t-1} + \eta_t \beta_t
	\nabla f_{\xi_t}(W^{t-1}), \quad \forall t > 0.
\]
The tentative iterate $\tilde W^t$ is then obtained by the proximal operation
associated with $\psi$:
\begin{equation}
\label{eq:daprox}
\tilde W^t = \prox_{\alpha_t\beta_t^{-1} \psi} \left( W^0
- \beta^{-1}_t V^t \right), \quad \alpha_t \coloneqq \sum\nolimits_{k=1}^t
\beta_k \eta_k,
\end{equation}
where for any function $g$, $\prox_g(x) \coloneqq \argmin_{y}\,
\norm{y-x}^2/2 + g(y)$ is its proximal operator.
The iterate is then updated along the direction $\tilde W^t - W^{t-1}$
with a factor of $c_t \in [0,1]$:
\begin{equation}
\label{eq:update}
	W^t = \left( 1 - c_t \right) W^{t-1} + c_t \tilde W^t = W^{t-1} + c_t
	\left( \tilde W^t - W^{t-1} \right).
\end{equation}
When $\psi \equiv 0$, \rmda reduces to the modernized dual averaging
algorithm of \citet{JS20a},
in which case it has been shown that mixing $W^{t-1}$ and
$\tilde W^t$ in \cref{eq:update} equals to introducing momentum
\citep{JS20a,tao2018primal}.
We found that this introduction of momentum greatly improves the
performance of \rmda and is therefore essential for applying it on
deep learning problems.

\begin{algorithm}[tb]
\LinesNumbered
\DontPrintSemicolon
\caption{\rmda$(W^0, T, \eta(\cdot), c(\cdot))$}
\label{alg:rmda}
\Input{Initial point $W^0$, learning rate schedule $\eta(\cdot)$, 
momentum schedule $c(\cdot)$, number of epochs $T$}

$V_0 \leftarrow 0,\quad \alpha_0 \leftarrow 0$

\For{$t=1,\dotsc,T$}
{
	$\beta_{t} \leftarrow \sqrt{t},
	\quad s_{t} \leftarrow \eta(t) \beta_{t},
	\quad \alpha_t \leftarrow \alpha_{t-1} + s_{t}$

	Sample $\xi_t \sim \gD$ and compute $V^t \leftarrow V^{t-1} +
	s_t \nabla f_{\xi_t}(W^{t-1})$

	$\tilde W^{t} \leftarrow \argmin_W\, \inprod{V^t}{W} +
	\frac{\beta_{t}}{2} \norm{W - W^0}^2 + \alpha_{t}\psi(W)$
	\Comment*[r]{\cref{eq:daprox}}
	\label{eq:prox}

	$W^t \leftarrow (1-c(t))W^{t-1}+ c(t) \tilde W^{t} $
}
\Output{The final model $W^T$}
\end{algorithm}

\section{Analysis}
\label{sec:analysis}
We provide theoretical analysis of the proposed \rmda in this section.
Our analysis shows variance reduction in \rmda and stationarity of
the limit point of its iterates, but all of them revolves around our
main purpose of identification of a stationary structure within a
finite number of iterations.
The key tools for this end are partial smoothness and manifold
identification \citep{HarL04a, Lew02a}.
Our result is the currently missing cornerstone for those proximal
algorithms applied to deep learning problems for identifying desired
structures.
In fact, it is actually well-known in convex optimization that those
algorithms based on plain proximal stochastic gradient without
variance reduction are \emph{unable to identify the active manifold},
and the structure of the iterates
oscillates due to the variance in the gradient estimation; see, for
example, experiments and discussions in \cite{LeeW12a,SunJNS19a}.
Our work is therefore the first one to provide justification for
solving the regularized optimization problem in deep learning to
really identify a desired structure induced by the regularizer.
Throughout, $\nabla f_\xi$ denotes the gradient of $f_\xi$,
$\partial \psi$ is the (regular) subdifferential of $\psi$,
and $\text{relint}(C)$ means the relative interior of the set $C$.

We start from introducing the notion of partial smoothness.
\begin{definition}
	\label{def:ps}
	A function $\psi$ is partly smooth at a point $W^*$ relative to a
	set $\gM_{W^*} \ni W^*$ if
	\begin{enumerate}[leftmargin=*]
		\itemsep 10pt
		\parskip -10pt
		\topsep -5pt
			\item Around $W^*$, $\gM_{W^*}$ is a $\mathcal{C}^2$-manifold and
				$\psi|_{\gM_{W^*}}$ is $\mathcal{C}^2$.
			\item $\psi$ is regular
 (finite with the Fr\'{e}chet subdifferential coincides with the
 limiting Fr\'{e}chet subdifferential) at all points $W \in \gM_{W^*}$
 around $W^*$ with $\partial \psi(W) \neq \emptyset$.
		\item The affine span of $\partial \psi(W^*)$ is a translate
			of the normal space to $\gM_{W^*}$ at $W^*$.
		\item $\partial \psi$ is continuous at $W^*$ relative to $\gM_{W^*}$.
	\end{enumerate}
\end{definition}
We often call $\gM_{W^*}$ the active manifold at $W^*$.
Another concept required for manifold identification is
prox-regularity \citep{RP96a}.
\begin{definition}
A function $\psi$ is prox-regular at $W^*$ for
$V^* \in \partial \psi(W^*)$ if $\psi$ is finite at $W^*$, locally lower
semi-continuous around $W^*$, and there is $\rho > 0$ such that
$\psi(W_1) \geq \psi(W_2) + \inprod{V}{W_1 - W_2} - \frac{\rho}{2}
\norm{W_1 - W_2}^2$ whenever $W_1,W_2$ are close to $W^*$ with
$\psi(W_2)$ near $\psi(W^*)$ and $V \in \partial \psi(W_2)$ near
$V^*$. $\psi$ is prox-regular at $W^*$ if it is so for all $V \in
\partial \psi(W^*)$.
\end{definition}
To broaden the applicable range,
a function $\psi$ prox-regular at some $W^*$ is often also assumed to be subdifferentially
continuous \citep{RP96a} there,
meaning that if $W^t \rightarrow W^*$, $\psi(W^t) \rightarrow
\psi(W^*)$ holds when there are $V^* \in \partial \psi (W^*)$ and a
sequence $\{V^t\}$ such that $V^t \in \partial \psi (W^t)$ and $V^t
\rightarrow V^*$.
Notably, all convex and weakly-convex \citep{EAN73a} functions are
regular, prox-regular, and subdifferentially continuous in their domain.

\subsection{Theoretical Results}
When the problem is convex, convergence guarantees for \cref{alg:rmda}
under two specific specific schemes are known.
First, when $c_t \equiv 1$, \rmda reduces to the classical RDA, and
convergence to a global optimum (of $W^t = \tilde W^t$ in this case)
on convex problems has been proven by \cite{LeeW12a,JD21a},
with convergence rates of the expected objective or the regret given
by \cite{LX10a,LeeW12a}.
Second, when $c_t = s_{t+1} \alpha_{t+1}^{-1}$ and $(\beta_t,
\alpha_t)$ in \cref{eq:prox} of \cref{alg:rmda} are replaced by
$(\beta_{t+1}, \alpha_{t+1})$,
convergence is recently analyzed by \citet{VK21a}.
In our analysis below, we do not assume convexity of either term.

We show that if $\{\tilde W^t\}$ converges to a point $W^*$
(which could be a non-stationary one), $\{W^t\}$ also converges to
$W^*$.
\begin{lemma}
\label{lemma:conv}
Consider \cref{alg:rmda} with $\{c_t\}$ satisfying
$\sum c_t = \infty$.
If $\{\tilde W^t\}$ converges to a point $W^*$, $\{W^t\}$ also
converges to $W^*$.
\end{lemma}
We then show that if $\{\tilde W^t\}$ converges to a point, almost
surely this point of convergence is stationary.
This requires the following lemma for variance reduction of \rmda,
meaning that the variance of using $V_t$ to estimate $\nabla
f(W^{t-1})$ reduces to zero, as $\alpha_t^{-1} V_t$ converges to
$\nabla f(W^{t-1})$ almost surely, and this result could be of its own
interest.
The first claim below uses a classical result in stochastic
optimization that can be found at, for example, \cite[Theorem~4.1,
Chapter~2.4]{AG79a}, but the second one is, to our knowledge, new.

\begin{lemma}
\label{lemma:iterate}
Consider \cref{alg:rmda}.
Assume for any $\xi \sim \gD$, $f_\xi$ is
$L$-Lipschitz-continuously-differentiable almost surely for some $L$,
so $f$ is also $L$-Lipschitz-continuously-differentiable,
and there is $C \geq 0$ such that
$\E_{\xi_t \sim \gD}\norm{\nabla f_{\xi_t}\left( W^{t-1}
\right)}^2 \leq C$ for all $t$.
If $\{\eta_t\}$ satisfies
\begin{equation}
	\sum \beta_t \eta_t \alpha_t^{-1} = \infty,\quad
	\sum \left(\beta_t  \eta_t \alpha_t^{-1}\right)^2 < \infty, \quad
	\norm{W^{t+1} - W^t}\left(\beta_t
	\eta_t\alpha_t^{-1}\right)^{-1} \as 0,
\label{eq:schedule}
\end{equation}
then $\alpha_t^{-1} V^t \longrightarrow \nabla f(W^{t-1})$ with
probability one.
Moreover, if $\{W^t\}$ lies in a bounded set, we get
$\E \norm{\alpha_t^{-1} V^t - \nabla
f\left( W^{t-1} \right)}^2 \rightarrow 0$ even if the second condition
in \cref{eq:schedule} is replaced by a weaker condition of $\beta_t
\eta_t \alpha_t^{-1} \rightarrow 0$.
\end{lemma}
In general, the last condition in \cref{eq:schedule} requires some
regularity conditions in $F$ to control the change speed of $W^t$.
One possibility is when $\psi$ is the indicator function of a convex
set, $\beta_t \eta_t \propto t^p$ for $t \in (1/2, 1)$ will
satisfy this condition.
However, in other settings for $\eta_t$, even when $F$ and $\psi$ are
both convex, existing analyses for the classical RDA such that $c_t
\equiv 1$ in \cref{alg:rmda} still need an additional local error
bound assumption to control the change of $W^{t+1} - W^t$.
Hence, to stay focused on our main message,  we take this assumption
for granted, and leave finding suitable sufficient conditions for it 
as future work.

With the help of \cref{lemma:iterate,lemma:conv}, we can now show the
stationarity result for the limit point of the iterates.
The assumption of $\beta_t \alpha_t^{-1}$ approaching $0$ below is
classical in analyses of dual averaging in order to gradually remove the
influence of the term $\norm{W - W^0}^2$.
\begin{theorem}
\label{thm:stationary}
Consider \cref{alg:rmda} with the conditions in
\cref{lemma:iterate,lemma:conv} hold,
and assume the set of stationary points $\gZ \coloneqq \{W \mid 0 \in
\partial F(W) \}$ is nonempty and $\beta_t \alpha_t^{-1} \rightarrow
0$.
For any given $W^0$, consider the event that $\{\tilde W^t\}$ converges to a
point $W^*$ (each event corresponds to a different $W^*$),
then if $\partial \psi$ is outer semicontinuous at $W^*$,
and this event has a nonzero probability, $W^* \in \gZ$, or
equivalently, $W^*$ is a stationary point, with probability one
conditional on this event.
\end{theorem}

Finally, with \cref{lemma:iterate,lemma:conv} and
\cref{thm:stationary}, we prove the main result that the active
manifold of the limit point is identified in finite iterations of
\rmda under nondegeneracy.
\begin{theorem}
Consider \cref{alg:rmda} with the conditions in \cref{thm:stationary}
satisfied.
Consider the event of $\{\tilde W^t\}$ converging to a certain point $W^*$
as in \cref{thm:stationary}, if the probability of this
event is nonzero; $\psi$ is prox-regular and
subdifferentially continuous at $W^*$ and partly smooth at $W^*$
relative to the active $\mathcal{C}^2$ manifold $\gM$; $\partial
\psi$ is outer semicontinuous at $W^*$; and the nondegeneracy condition
\begin{equation}
	-\nabla f\left( W^* \right) \in \textup{relint}\; \partial \psi
	\left( W^* \right)
	\label{eq:nod}
\end{equation}
holds at $W^*$,
then conditional on this event, almost surely there is $T_0 \geq 0$
such that
\begin{equation}
	\tilde W^t \in \gM,\quad \forall t \geq T_0.
	\label{eq:identify}
\end{equation}
In other words, the active manifold at $W^*$ is identified by the
iterates of \cref{alg:rmda} after a finite number of iterations almost
surely.
\label{thm:identify}
\end{theorem}

As mentioned in \cref{sec:intro}, an important reason for studying
manifold identification is to get the lowest-dimensional manifold
representing the structure of the limit point, which often corresponds
to a preferred property for the application, like the highest
sparsity, lowest rank, or lowest VC dimension locally. See
an illustrated example in \cref{app:sparse}.

\section{Applications in Deep Learning}
\label{sec:app}
We discuss two popular schemes of training structured deep learning models
achieved through regularization to demonstrate the applications of \rmda.
More technical details for applying our theory to the regularizers in
these applications are in \cref{app:applications}.

\subsection{Structured Sparsity}
\label{sec:sparse}

As modern deep NN models are often gigantic,
it is sometimes desirable to trim the model to a smaller one when only
limited resources are available.
In this case, zeroing out redundant parameters during training at the
group level is shown to be useful \citep{zhou2016less}, and one can
utilize regularizers promoting structured sparsity for this purpose.
The most famous regularizer of this kind is the group-LASSO norm
\citep{yuan2006model, friedman2010note}.
Given $\lambda \ge 0$ and a collection $\gG$ of index
sets $\{\gI_g\}$ of the variable $W$,
this convex regularizer is defined as
\begin{equation}
	\psi(W) \coloneqq \lambda \sum\nolimits_{g=1}^{|\gG|} w_g
	\norm{W_{\gI_g}},
\label{eq:grouplasso}
\end{equation}
with $w_g > 0$ being the pre-specified weight for $\gI_g$.
For any $W^*$, let $\gG_{W^*} \subseteq \gG$ be the index
set such that $W^*_{\gI_j} = 0$ for all $j \in \gG_{W^*}$,
the group-LASSO norm is partly smooth around $W^*$ relative to the
manifold $\gM_{W^*} \coloneqq \{W \mid W_{\gI_i} = 0, \forall i \in
\gG_{W^*} \}$,
so our theory applies.


In order to promote structured sparsity, we need to carefully design
the grouping.
Fortunately, in NNs, the parameters can be grouped naturally
\citep{wen2016learning}.
For any fully-connected layer,
let $W \in \R^{\text{out} \times \text{in}}$ be the matrix
representation of the associated parameters, where $\text{out}$ is the
number of output neurons and $\text{in}$ is that of input neurons,
we can consider the column-wise groups, defined as ${W}_{:,j}$ for all
$j$, and the row-wise groups of the form ${W}_{i,:}$.
For a convolutional layer with $W \in \R^{\text{filter} \times
\text{channel} \times \text{height} \times \text{width}}$ being the
tensor form of the corresponding parameters,
we can consider channel-wise, filter-wise, and kernel-wise groups, defined
respectively as ${W}_{:,j,:,:}$, $W_{i,:,:,:}$ and $W_{i,j,:,:}$.

\subsection{Binary/Discrete Neural Networks}
Making the parameters of an NN binary integers is another
way to obtain a more compact model during training and deployment
\citep{hubara2016binarized}, but discrete optimization is hard to
scale-up.
Using a vector representation $w \in \R^m$ of the variables,
\citet{hou2016loss}
thus proposed to use the indicator function of $\left\{ w \mid
	w_{\gI_i} = \alpha_i b_{\gI_i}, \alpha_i > 0, b_{\gI_i} \in \{\pm
	1\}^{|\gI_i|}  \right\}$ to induce the entries of $w$ to be binary
	without resorting to discrete optimization tools, where each
	$\gI_i$ enumerates all parameters in the $i$-th layer.
\citet{yang2019proxsgd} later proposed to use
$\min\nolimits_{\alpha \in [0, 1]^{m}} \quad \sum\nolimits_{i=1}^{m}
\left(\alpha_i(w_i
+ 1)^2 + (1 - \alpha_i)(w_i - 1)^2\right)$
as the regularizer and to include $\alpha$ as a variable to train.
At any $\alpha^*$ with $I_0 \coloneqq \{i \mid \alpha_i^* =
0\}$ and $I_1 \coloneqq \{i \mid \alpha_i^* = 1\}$, the objective is
partly smooth relative to the manifold $\{(W,\alpha) \mid
	\alpha_{I_0} = 0, \alpha_{I_1} = 1\}$.
Extension to discrete NNs beyond the binary ones is
possible, and \citet{bai2018proxquant} have proposed regularizers
with closed-form proximal operators for it.

\section{Experiments}
\label{sec:exp}
We use the structured sparsity application in \cref{sec:sparse} to
empirically exemplify the ability of \rmda to find desired
structures in the trained NNs.
\rmda and the following methods for structured sparsity in deep
learning are compared using PyTorch \citep{AP19a}.
\vspace{-8pt}
\begin{itemize}[leftmargin=*]
\parskip 0pt
\topsep 0pt
\partopsep 0pt
\itemsep 0pt
\item \sgd \citep{yang2019proxsgd}: A simple proxMSGD algorithm.
	To obtain group sparsity, we skip the
	interpolating step in \cite{yang2019proxsgd}.
\item \ssi \citep{deleu2021structured}: This is a special case
	of the adaptive proximal SGD framework of
	\cite{JY20a} that uses the Newton-Raphson algorithm
	to approximately solve the subproblem.
	We directly use the package released by the
	authors.
\end{itemize}
\vspace{-8pt}
We exclude the algorithm of \citet{wen2016learning}
because their method is shown to be worse than \ssi by
\citet{deleu2021structured}.

To compare these algorithms, we examine both the validation
accuracy and the group sparsity level of their trained models.
We compute the group sparsity as the percentage of groups whose
elements are all zero, so the reported group sparsity is zero when
there is no group with a zero norm, and is one when the whole model is
zero.
For all methods above, we use \cref{eq:grouplasso} with column-wise
and channel-wise groupings in the regularization for training, but
adopt the kernel-wise grouping in their group sparsity evaluation.
Throughout the experiments, we always use multi-step learning rate
scheduling that decays the learning rate by a constant factor every
time the epoch count reaches a pre-specified threshold.
For all methods, we conduct grid searches to find the best
hyperparameters.
All results shown in tables in \cref{sec:correct,sec:NN} are the mean
and standard deviation of three independent runs with the same
hyperparameters, while figures use one representative run for better
visualization.

In convex optimization, a popular way to improve the practical
convergence behavior for momentum-based methods is restarting that
periodically reset the momentum to zero \citep{BO15a}.
Following this idea, we introduce a restart heuristic to \rmda.
At each round, we use the output of \cref{alg:rmda} from the previous
round as the new input to the same algorithm, and continue using the
scheduling $\eta$ and $c$ without resetting them.
For $\psi \equiv 0$, \citet{JS20a} suggested to increase $c_t$
proportional to the decrease of $\eta_t$ until reaching $c_t =1$.
We adopt the same setting for $c_t$ and $\eta_t$ and restart \rmda
whenever $\eta_t$ changes.
As shown in \cref{sec:analysis} that $\tilde W^t$ finds the
active manifold, increasing $c_t$ to $1$ also accords with our
interest in identifying the stationary structure.


\subsection{Correctness of Identified Structure Using Synthetic Data}
\label{sec:correct}
\ifdefined\arxiv
\def\len{1.4in}
\else
\def\len{.9in}
\fi
\begin{figure}[tbh]
\begin{center}
\begin{subfigure}[b]{0.32\textwidth}
\centering
\includegraphics[height=\len]{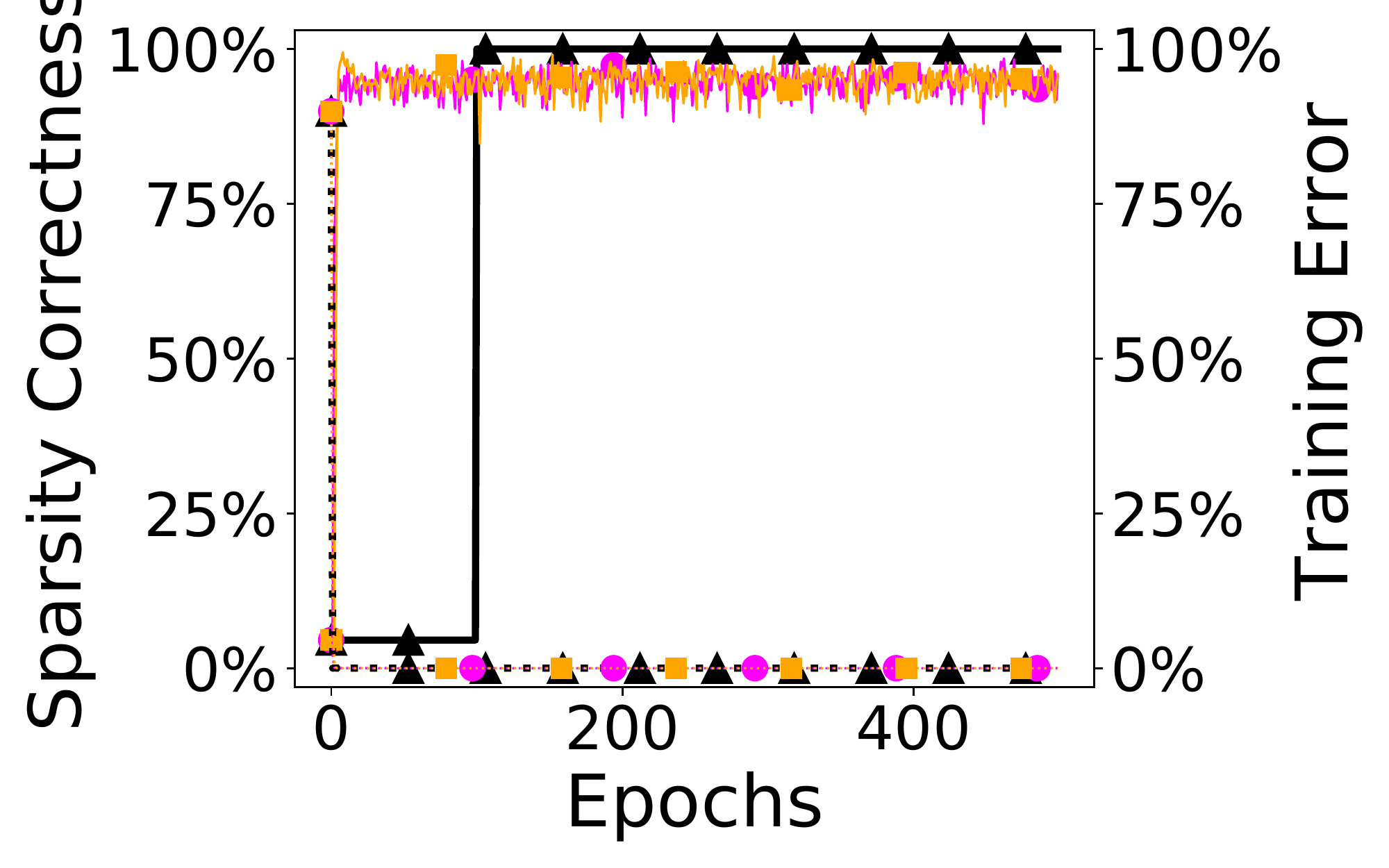}
\caption{Logistic regression}
\end{subfigure}
\begin{subfigure}[b]{0.32\textwidth}
\centering
\includegraphics[height=\len]{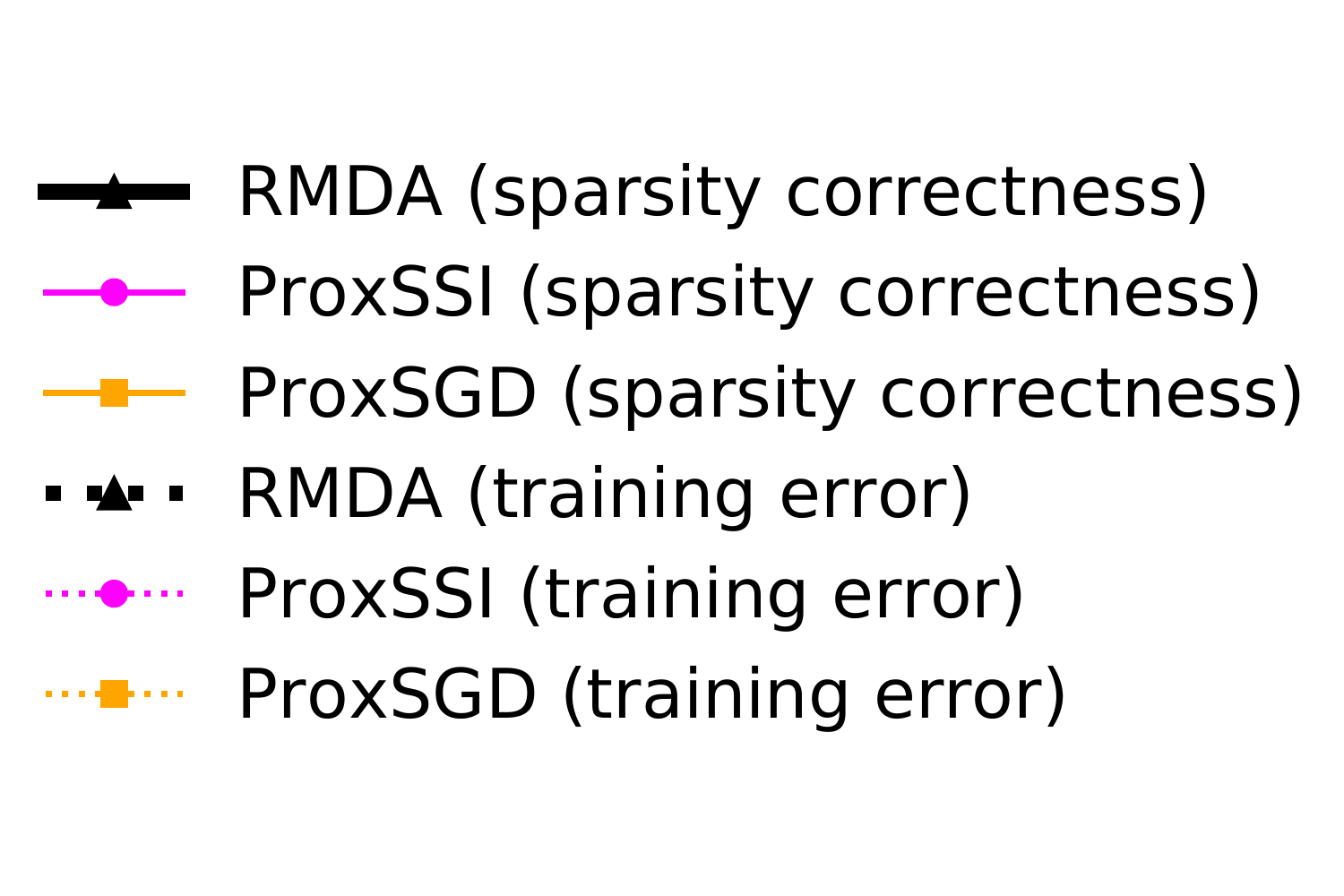}
\caption{Legend}
\end{subfigure}
\begin{subfigure}[b]{0.32\textwidth}
\centering
\includegraphics[height=\len]{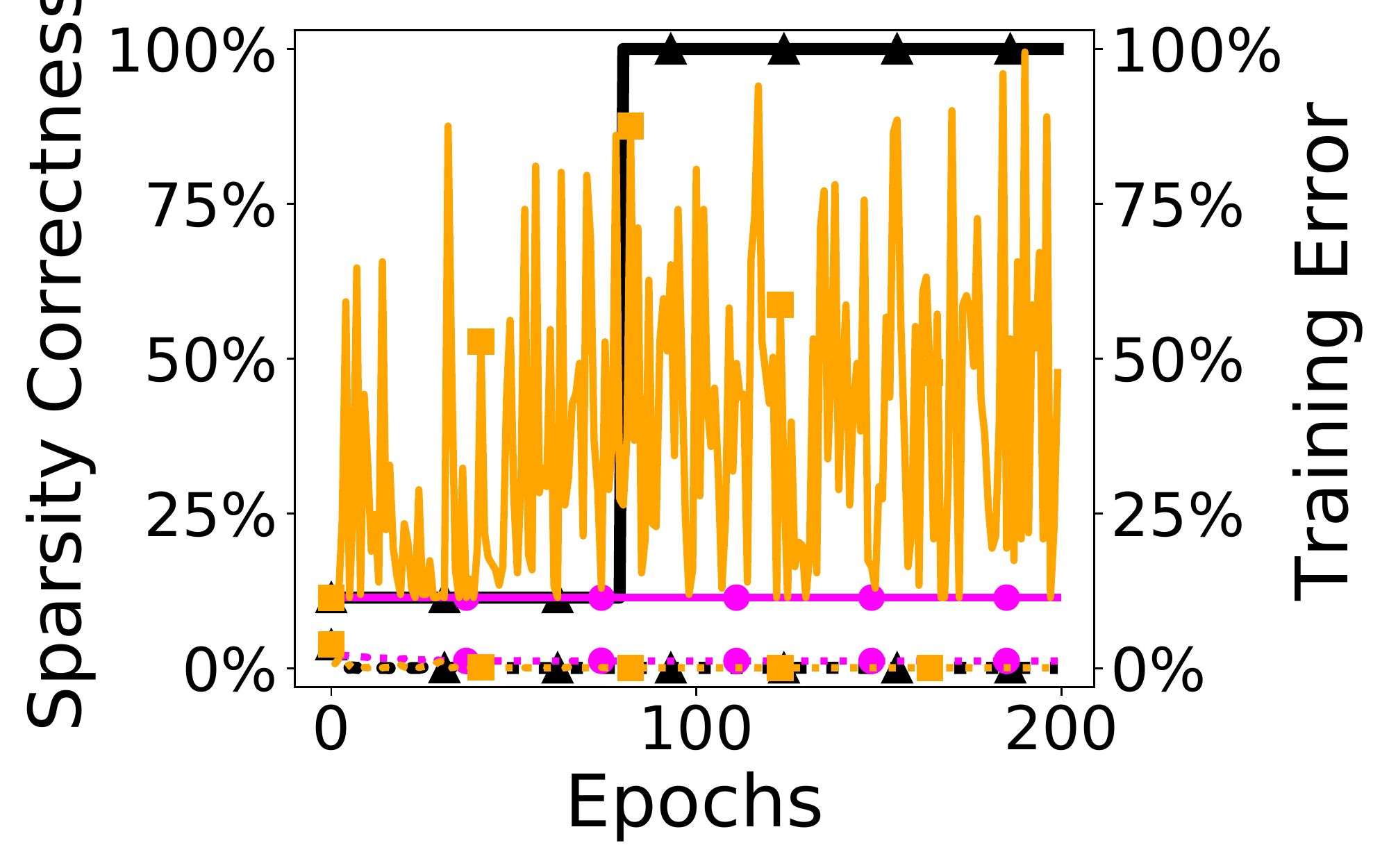}
\caption{Convolutional network}
\end{subfigure}
\end{center}
\caption{Group sparsity pattern correctness and training
	error rates on synthetic data.}
\label{fig:correct}
\end{figure}

Our first step is to numerically verify that \rmda can indeed identify
the stationary structure desired.
To exactly find a stationary point and its structure a priori, we
consider synthetic problems.
We first decide a ground truth model $W$ that is structured sparse,
generate random data points that can be well separated by $W$, and
then decide their labels using $W$.  The generated data are then taken
as our training data.
We consider a linear logistic regression model and a small NN that has
one fully-connected layer and one convolutional layer.
To ensure convergence to the ground truth,
for logistic regression we generate more data points than the problem
dimension to ensure the problem is strongly convex so that there is
only one stationary/optimal point,
and for the small NN, we initialize all algorithms close enough to the
ground truth.
We report in \cref{fig:correct} training error rates (as an indicator
for the proximity to the ground truth) and percentages of the optimal
group sparsity pattern of the ground truth identified.
Clearly, although all methods converge to the ground truth, only \rmda
identifies the correct structure of it, and other methods without
guarantees for manifold identification fail.

\subsection{Neural Networks with Real Data}
\label{sec:NN}
\begin{table}[tbh]
\renewcommand{\arraystretch}{0.3}
\caption{Group sparsity and validation accuracy of different methods.
We report mean and standard deviation of three independent runs
(except that for the linear convex model, only one run is conducted as
we are guaranteed to find the global optima).
MSGD is the baseline with no sparsity-inducing regularizer.}
\label{tbl:results_real}
\begin{center}
\sisetup{detect-weight,mode=text}
\renewrobustcmd{\bfseries}{\fontseries{b}\selectfont}
\renewrobustcmd{\boldmath}{}
\newrobustcmd{\B}{\bfseries}
\begin{center}
\begin{tabular}{@{}lrr|rr@{}}
Algorithm & Validation accuracy & Group sparsity                  & Validation accuracy & Group sparsity\\
\hline\\
& \multicolumn{2}{c|}{Logistic regression/MNIST} &
\multicolumn{2}{c}{Fully-connected NN/FashionMNIST}\\
\hline
\sgd & 91.31 \% & 38.78 \%		& 88.72 $\pm$ 0.05\% & 31.42 $\pm$ 0.36\% \\
\ssi & 91.31  \% & 39.54 \%		& 88.44 $\pm$ 0.41\% & 35.25 $\pm$ 1.56\%\\
\rmda & \B 91.34 \% & \B 56.51 \% & \B 88.09 $\pm$ 0.04\% & \B {42.89 $\pm$ 0.66\%}\\
\hline\\
& \multicolumn{2}{c|}{LeNet5/MNIST} &
\multicolumn{2}{c}{LeNet5/FashionMNIST}\\
\hline
MSGD  & 99.36 $\pm$ 0.06\% & \multicolumn{1}{c|}{-} & 91.96 $\pm$ 0.01\% & \multicolumn{1}{c}{-} \\
\hline
\sgd & \B 99.13 $\pm$ 0.02\% & 76.57 $\pm$ 2.33\%             & 90.99 $\pm$ 0.17\% & 50.50 $\pm$ 2.66\% \\
\ssi & 99.07 $\pm$ 0.03\% & 77.82 $\pm$ 1.56\%              & 90.93 $\pm$ 0.02\% & 60.49 $\pm$ 1.05\% \\
\rmda & 99.10 $\pm$ 0.06\% & \B{79.81 $\pm$ 1.56\%}         & \B 91.41 $\pm$ 0.10\% & \B{66.15 $\pm$ 1.68\%} \\
\hline\\
& \multicolumn{2}{c|}{VGG19/CIFAR10} &
\multicolumn{2}{c}{VGG19/CIFAR100}\\
\hline
MSGD  & 94.03 $\pm$ 0.11\% & \multicolumn{1}{c|}{-} & 74.62 $\pm$ 0.22\% & \multicolumn{1}{c}{-} \\
\hline
\sgd & 92.38 $\pm$ 0.31\% & 72.57 $\pm$ 6.04\%             & 71.91 $\pm$ 0.08\% & 08.63 $\pm$ 4.88\% \\
\ssi & 92.51 $\pm$ 0.03\% & 81.05 $\pm$ 0.16\%              & 66.20 $\pm$ 0.38\% & 46.41 $\pm$ 1.42\% \\
\rmda & \B 93.62 $\pm$ 0.15\% & \B{86.37 $\pm$ 0.25\%}         & \B 72.23 $\pm$ 0.20\% & \B{58.86 $\pm$ 0.41\%} \\
\hline\\
& \multicolumn{2}{c|}{ResNet50/CIFAR10} &
\multicolumn{2}{c}{ResNet50/CIFAR100}\\
\hline
MSGD  & 95.65 $\pm$ 0.03\% & \multicolumn{1}{c|}{-} & 79.13 $\pm$ 0.19\% & \multicolumn{1}{c}{-} \\
\hline
\sgd & 92.36 $\pm$ 0.14\% & 76.82 $\pm$ 4.09\%             & 75.53 $\pm$ 0.49\% & 51.83 $\pm$ 0.34\% \\
\ssi & 94.09 $\pm$ 0.08\% & 74.81 $\pm$ 1.28\%              & 74.52 $\pm$ 0.29\% & 32.79 $\pm$ 2.53\% \\
\rmda & \B 94.25 $\pm$ 0.02\% & \B{83.01 $\pm$ 0.50\%}         & \B 76.12 $\pm$ 0.46\% & \B{57.67 $\pm$ 3.76\%} \\
\end{tabular}
\end{center}
\end{center}
\end{table}

\begin{figure}[tb]
	\centering
	\begin{tabular}{@{}c@{}c@{}c@{}c@{}}
	    \includegraphics[width=.25\textwidth]{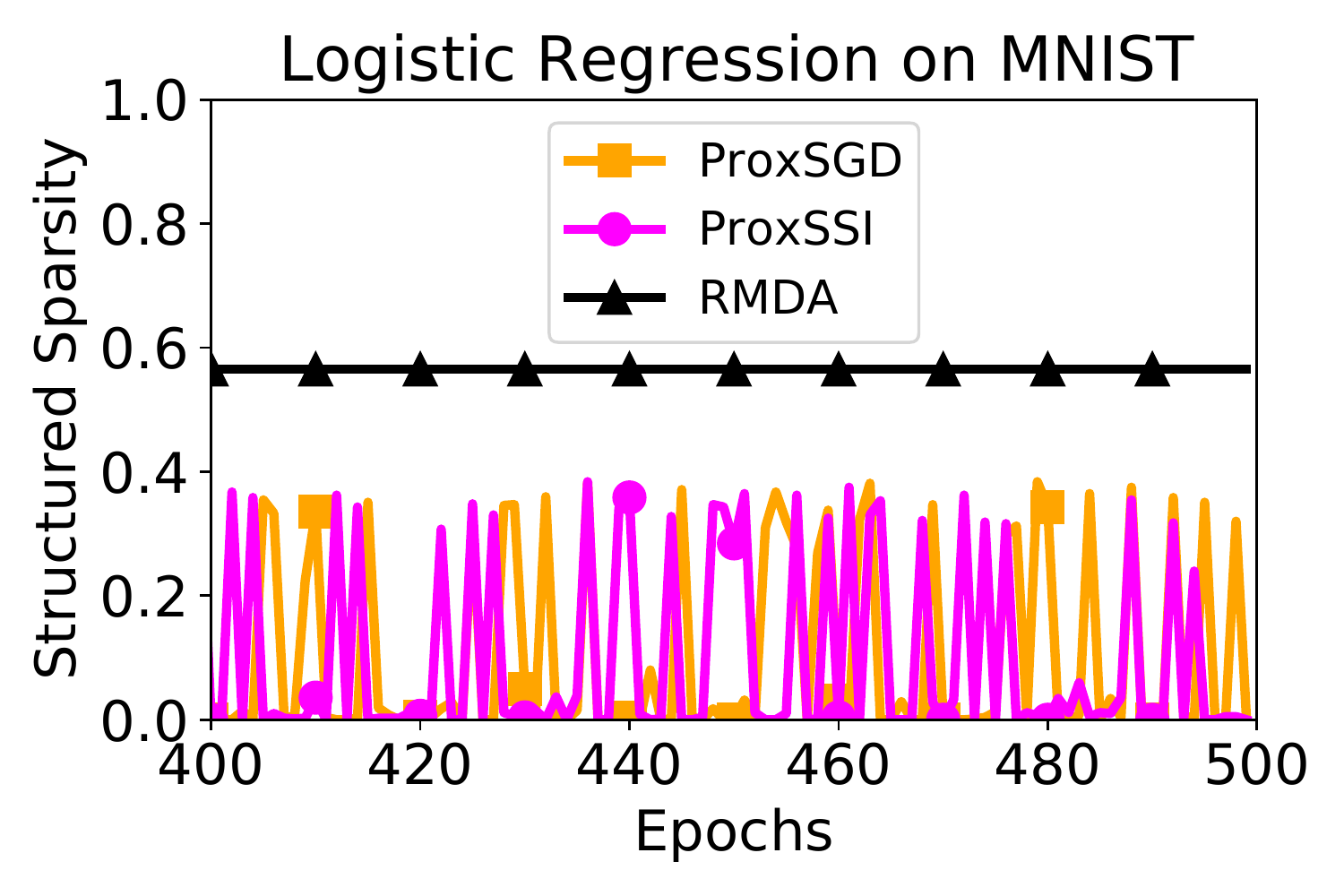}&
		\includegraphics[width=.25\textwidth]{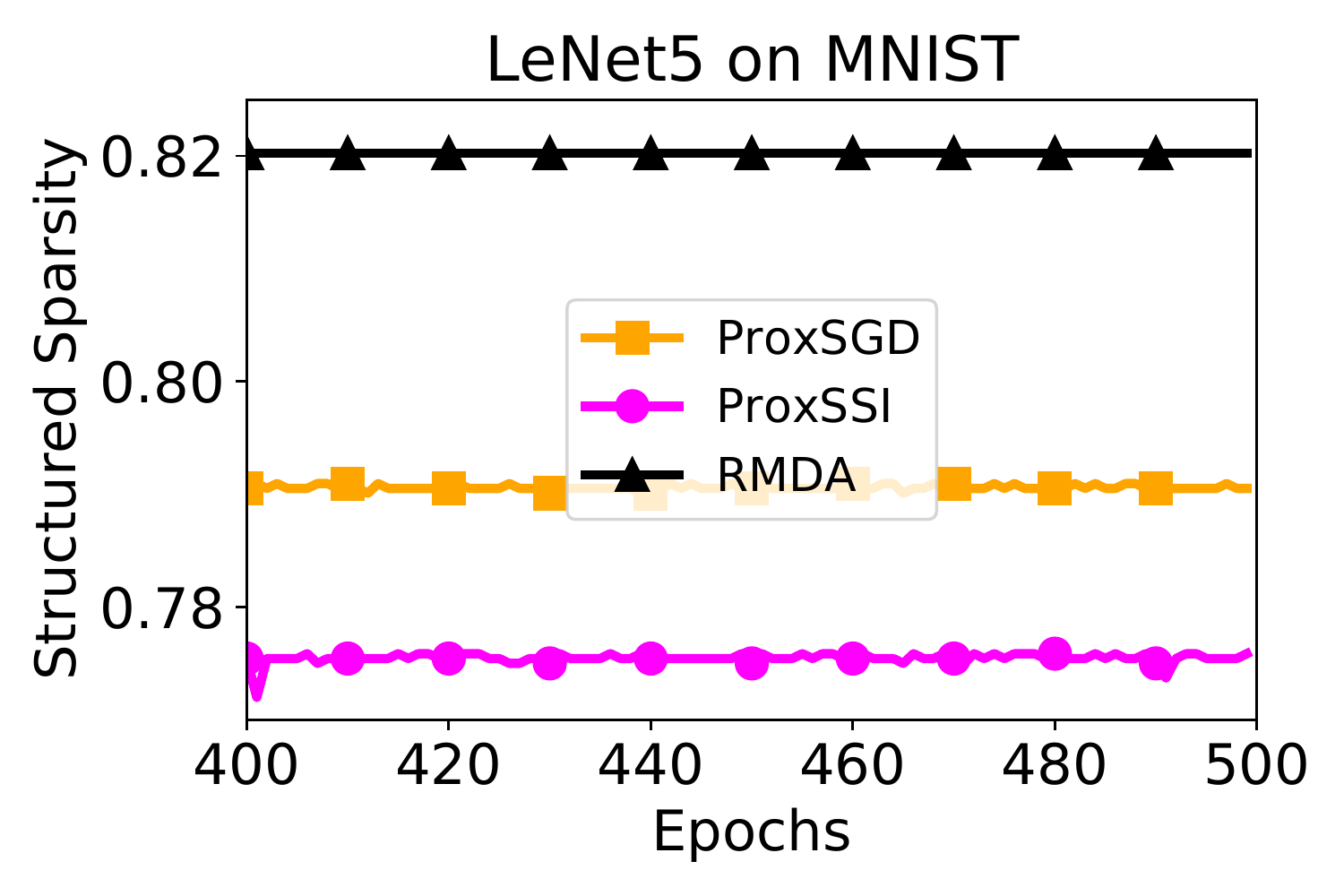}&
		\includegraphics[width=.25\textwidth]{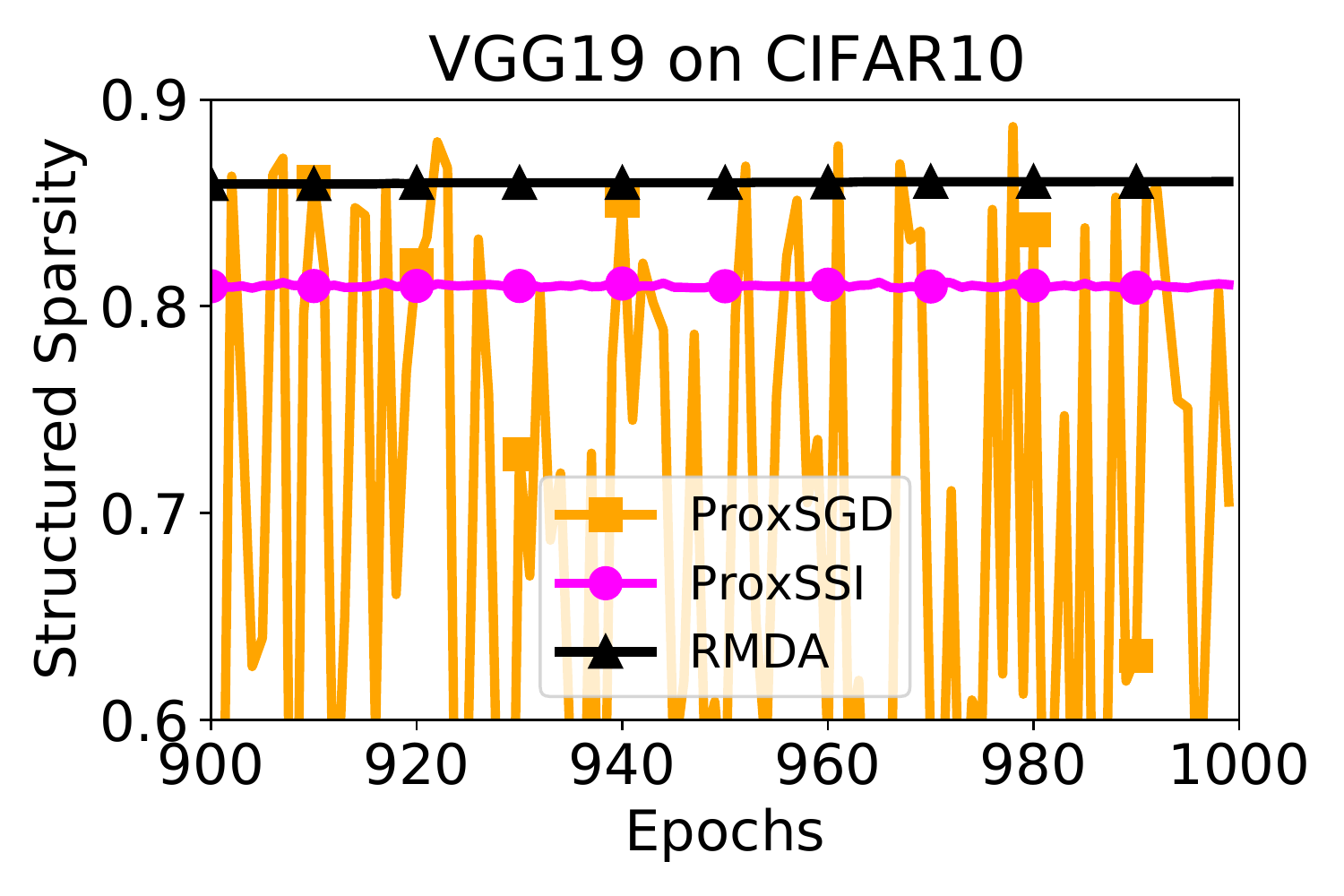}&
		\includegraphics[width=.25\textwidth]{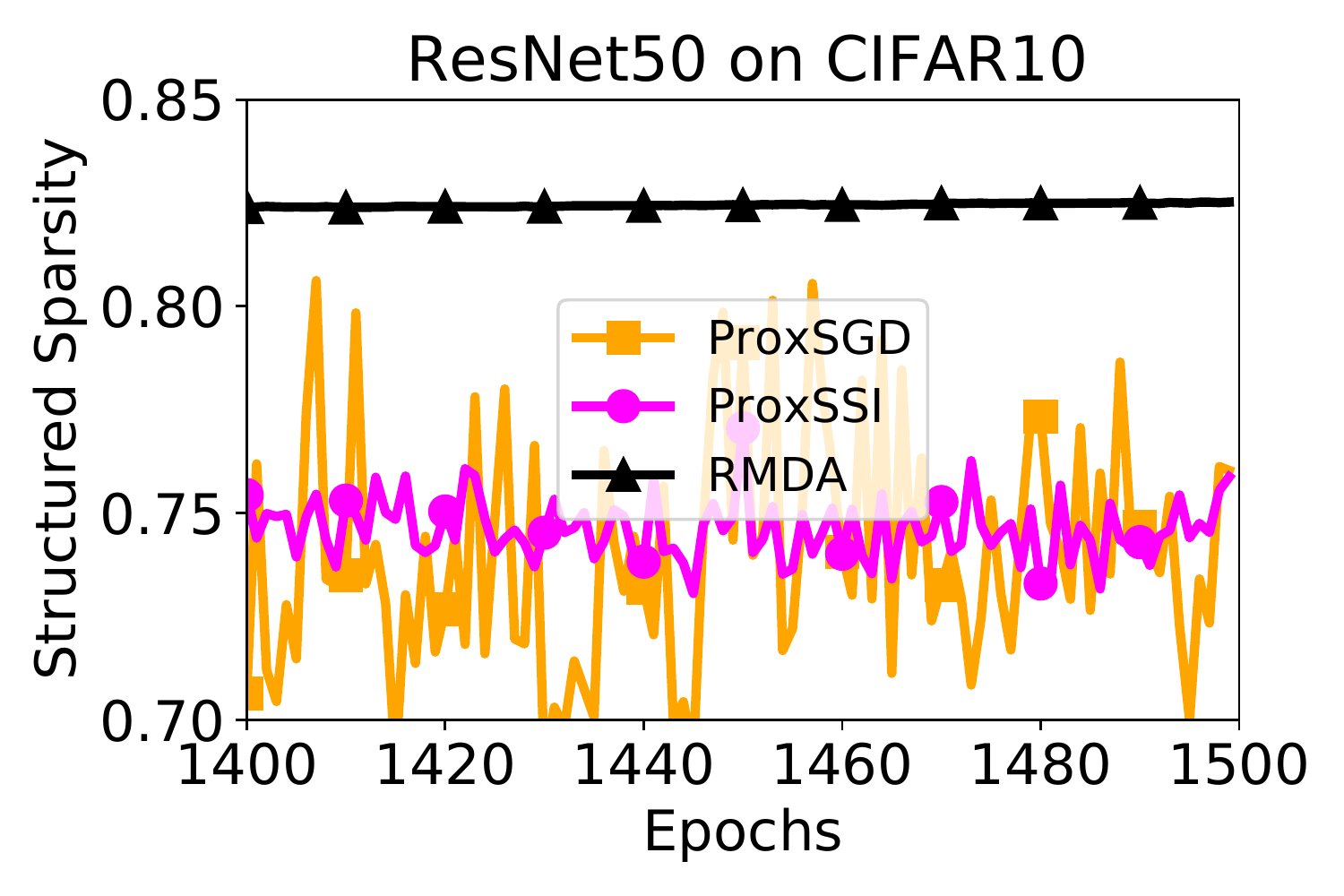}\\
		\includegraphics[width=.25\textwidth]{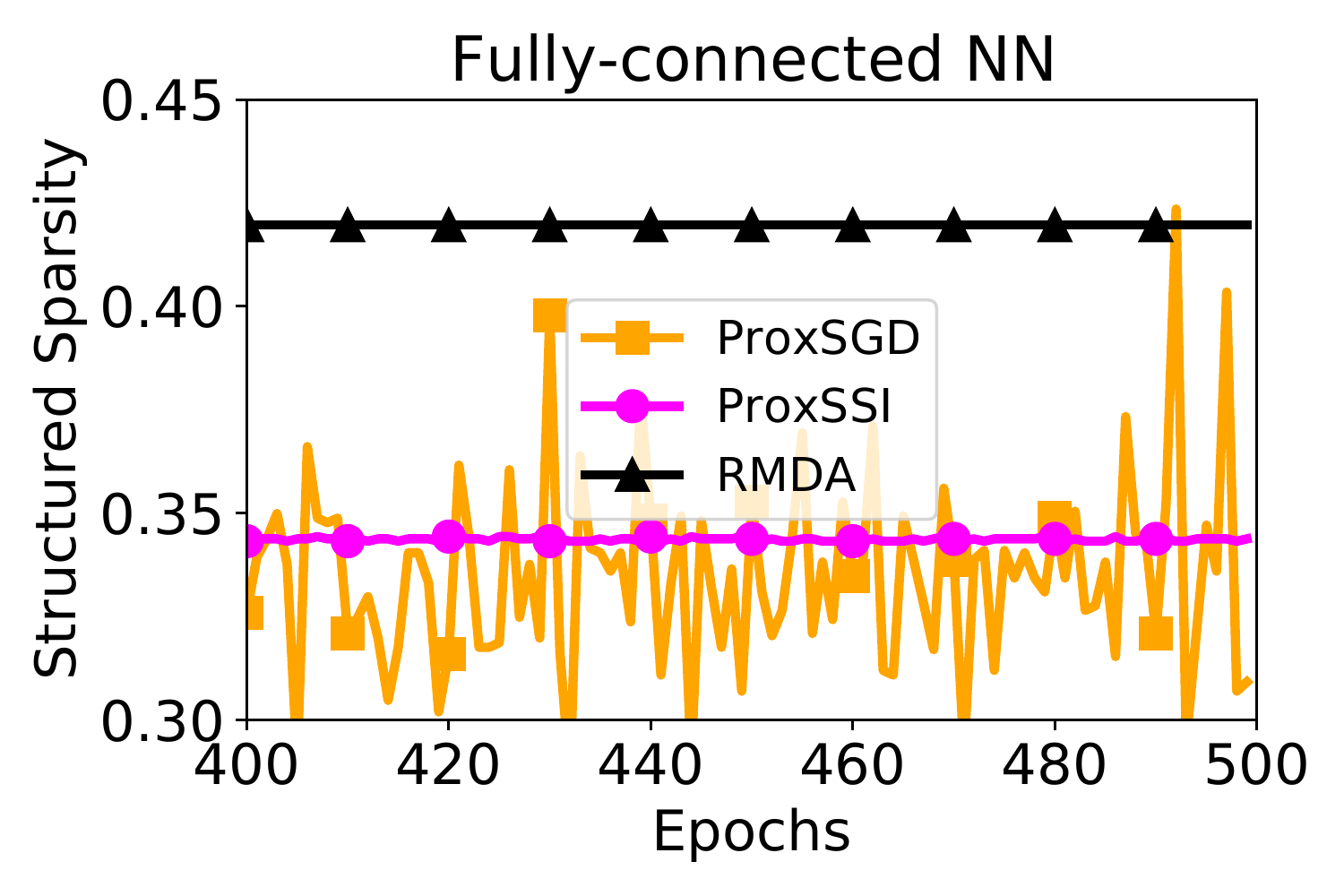}&
		\includegraphics[width=.25\textwidth]{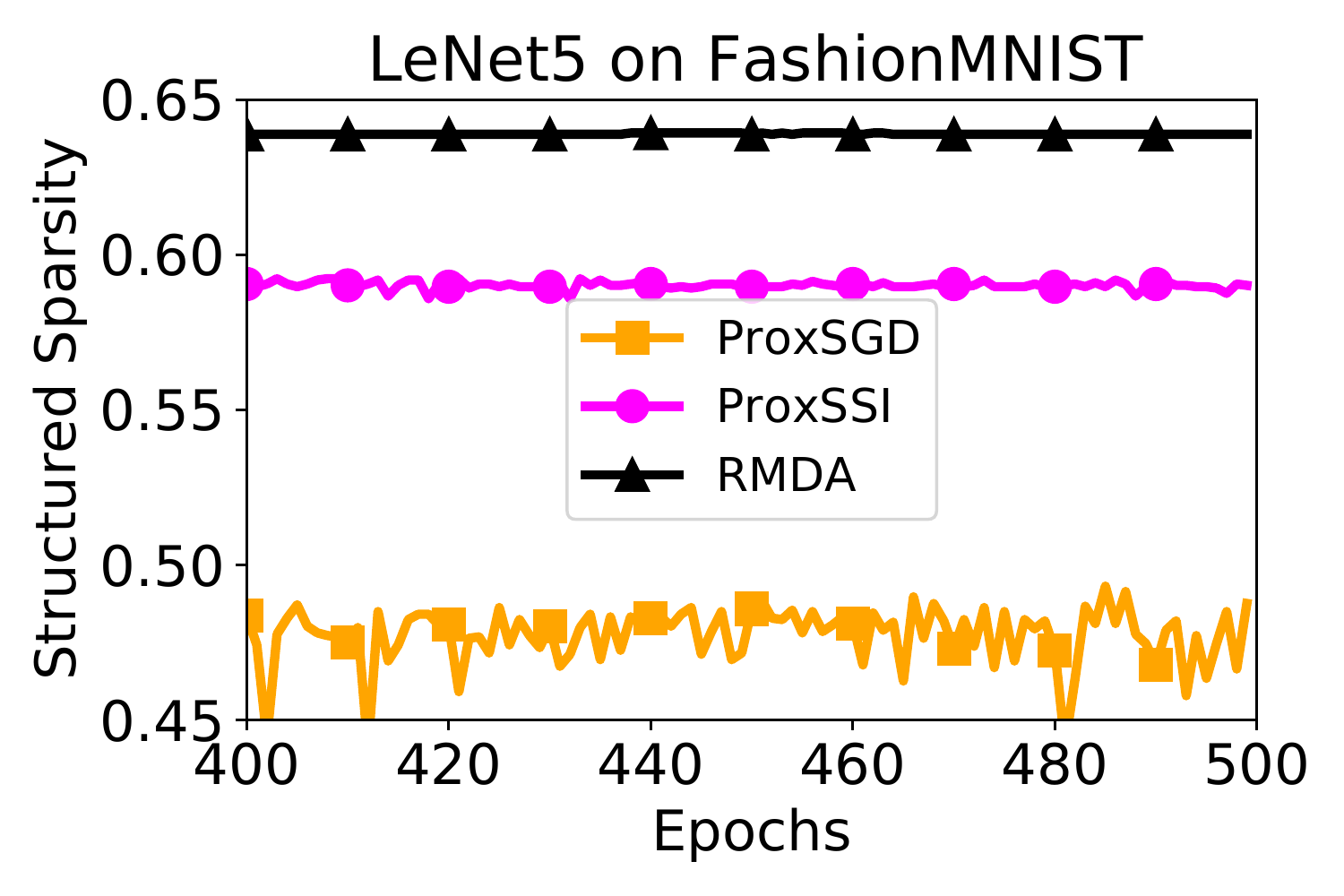}&
		\includegraphics[width=.25\textwidth]{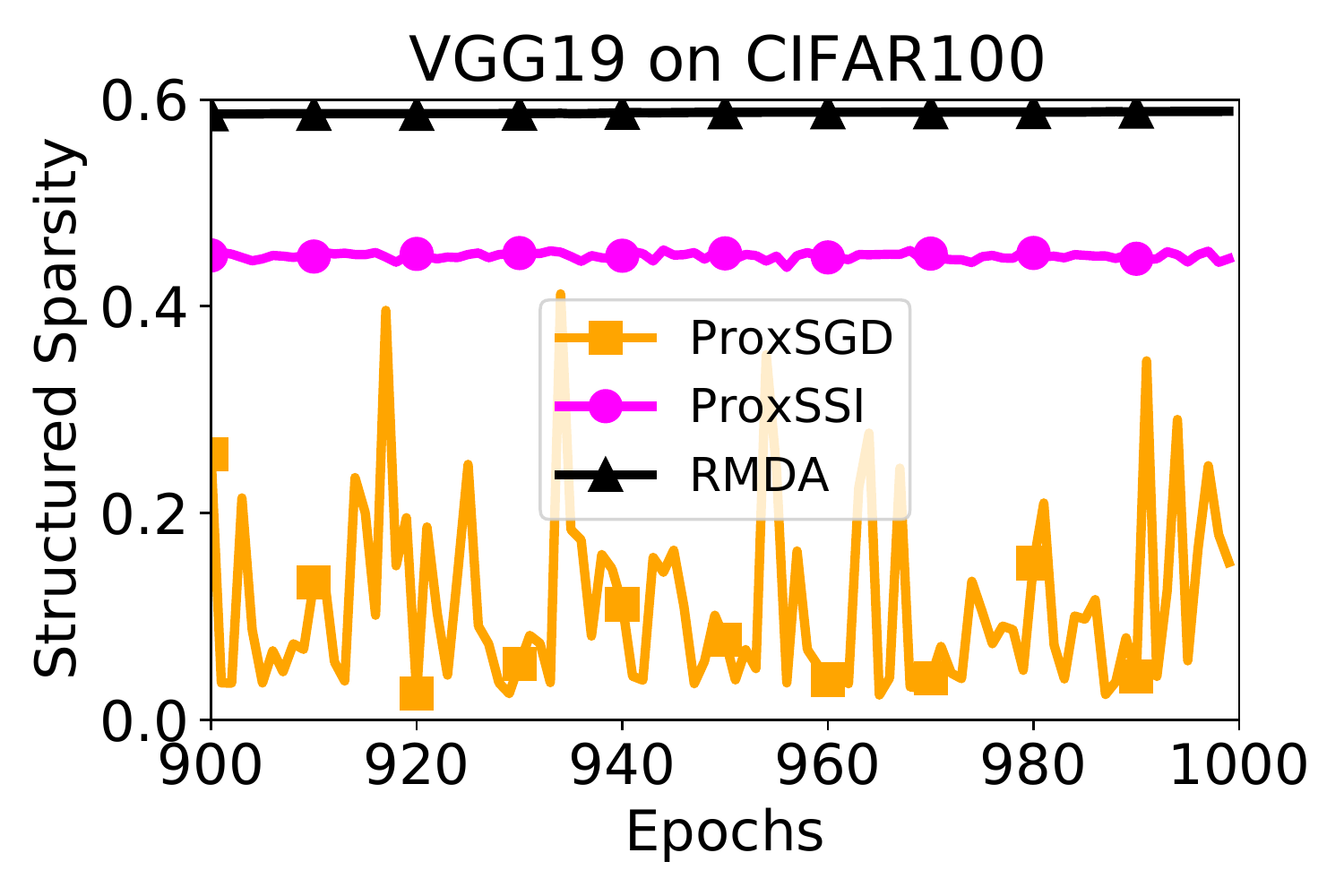}&
		\includegraphics[width=.25\textwidth]{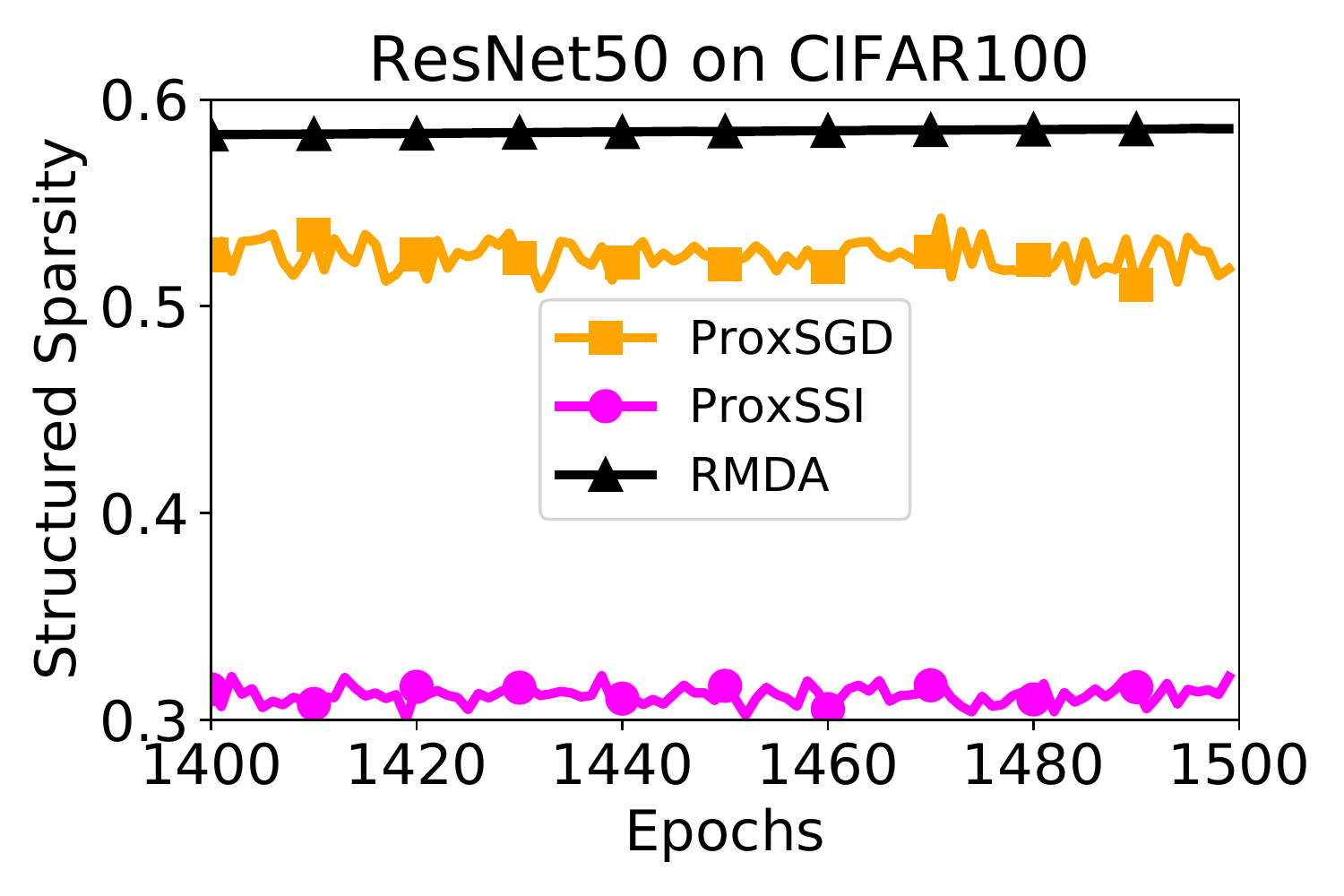}
	\end{tabular}	
\caption{Group Sparsity v.s epochs of different algorithms on NNs of a
single run.}
 \label{fig:NN_sparsity}
\end{figure}

We turn to real-world data used in modern computer vision problems.
We consider two rather simple models and six more complicated modern
CNN cases.
The two simpler models are linear logistic regression with the MNIST dataset
\citep{lecun1998gradient}, and training a small NN with seven fully-connected
layers on the FashionMNIST dataset \citep{xiao2017fashion}.
The six more complicated cases are:
\begin{enumerate}
		\itemsep 10pt
		\parskip -10pt
		\topsep -5pt
	\item A version of LeNet5 with the MNIST dataset,
	\item The same version of LeNet5 with the FashionMNIST dataset,
	\item A modified VGG19 \citep{simonyan2014very} with the CIFAR10
		dataset \citep{krizhevsky2009learning},
	\item The same modified VGG19 with the CIFAR100 dataset
		\citep{krizhevsky2009learning},
	\item ResNet50 \citep{he2016deep}  with the CIFAR10 dataset, and
	\item ResNet50  with the CIFAR100 dataset.
\end{enumerate}
For these six more complicated tasks, we include a dense baseline of MSGD
with no sparsity-inducing regularizer in our comparison.
For all training algorithms on VGG19 and ResNet50, we follow the
standard practice in modern vision tasks to apply data augmentation
through random cropping and horizontal flipping so that the training
problem is no longer a finite-sum one.
From \cref{fig:NN_sparsity}, we see that similar to
the previous experiment, the group sparsity level of \rmda is stable in
the last epochs, while that of \sgd and \ssi oscillates below.
This suggests that \rmda is the only method that, as proven in
\cref{sec:analysis}, identifies the structured sparsity at its limit
point, and other methods with no variance reduction fail.
Moreover, \cref{tbl:results_real} shows that manifold identification
of \rmda is achieved with no sacrifice of the validation accuracy,
so \rmda beats \sgd and \ssi in both criteria, and its accuracy is
close to that of the dense baseline of MSGD.
Moreover, for VGG19 and ResNet50, \rmda succeeds in finding the
optimal structured sparsity pattern despite the presence of data augmentation,
showing that \rmda can indeed overcome the difficulty from the
infinite-sum setting of modern deep learning tasks.

We also report that in the ResNet50/CIFAR100 task, on our NVIDIA RTX 8000
GPU, MSGD, \sgd, and \rmda have similar per-epoch cost of $68$,
$77$, and $91$ seconds respectively,
while \ssi needs $674$ seconds per epoch.
\rmda is thus also more suitable for large-scale structured deep
learning in terms of practical efficiency.

\ifdefined\arxiv
\def\len{1.3in}
\else
\def\len{.92in}
\fi

\subsection{Comparison with Pruning}
\label{sec:cp}
We compare \rmda with a state-of-the-art pruning method \rigl
\citep{UE20a}.
As pruning focuses on unstructured sparsity, we
use \rmda with $\psi(W) = \lambda \norm{W}_1$ to have a fair
comparison, and tune $\lambda$ to achieve a pre-specified
sparsity level.
We run \rigl with $1000$ epochs, as its performance at the default 500
epochs was unstable, and let \rmda use the same number of epochs.
Results of $98\%$ sparsity in \cref{tbl:prune} show that
\rmda consistently outdoes \rigl,
indicating regularized training could be a promising
alternative to pruning.

\begin{table}[tbh]
\renewcommand{\arraystretch}{0.3}
\caption{Comparison between \rmda and \rigl with $1000$ epochs for
unstructured sparsity in a single run.}
\label{tbl:prune}
\sisetup{detect-weight,mode=text}
\renewrobustcmd{\bfseries}{\fontseries{b}\selectfont}
\renewrobustcmd{\boldmath}{}
\newrobustcmd{\B}{\bfseries}
\centering
\begin{tabular}{lrr|rr}
	& \multicolumn{2}{c|}{ResNet50 with CIFAR10} &
	\multicolumn{2}{c}{ResNet50 with CIFAR100}\\
	\hline
	Algorithm & Sparsity & Accuracy & Sparsity & Accuracy\\
	\hline
	Dense baseline & & 94.81\% & &  74.61\%\\
	\hline
	\rmda & \B 98.36\% & \B 93.78\% & \B 98.32\% & \B 74.32\%\\
	\rigl & 98.00\% & 93.41\% & 98.00\% & 70.88\%
\end{tabular}
\end{table}
\section{Conclusions}
In this work, we proposed and analyzed a new algorithm, \rmda, for efficiently
training structured neural networks with state-of-the-art performance.
Even in the presence of data augmentation, \rmda
can still achieve variance reduction and provably identify the desired
structure at a stationary point using the tools of manifold identification.
Experiments show that existing algorithms for the same purpose fail to
find a stable stationary structure, while \rmda achieves so with no
accuracy drop nor additional time cost.

\ifdefined\arxiv
\else
\pagebreak
\fi
\section*{Acknowledgements}
This work was supported in part by MOST of R.O.C. grant
109-2222-E-001-003-MY3, and the AWS Cloud Credits for Research program
of Amazon Inc.

\bibliographystyle{iclr2022_conference}
\bibliography{rmdasparse}

\appendix
\part{Appendices} 
\parttoc 
\section{Proofs}
\label{sec:proof}
\subsection{Proof of \cref{lemma:conv}}
\begin{proof}
Using \cref{eq:update}, the distance between $W^t$ and $W^*$ can be
upper bounded through the triangle inequality:
\begin{align}
\nonumber
\norm{W^t - W^*} &=
\norm{\left( 1 - c_t \right) \left( W^{t-1} - W^* \right) + c_t \left(
\tilde W^t - W^* \right)}\\
&\leq c_t \norm{\tilde W^t - W^*} + \left( 1 - c_t \right)
\norm{W^{t-1} - W^*}.
\label{eq:upper}
\end{align}
For any event such that $\tilde W^t \rightarrow W^*$, for any
$\epsilon > 0$, we can find $T_\epsilon \geq 0$ such that
\[
	\norm{\tilde W^t - W^*} \leq \epsilon,\quad \forall t \geq T_\epsilon.
\]
Let $\delta_t \coloneqq \norm{W^t - W^*}$, we see from the above
inequality and \cref{eq:upper} that
\begin{equation*}
\delta_t \leq \left( 1 - c_t \right)\delta_{t-1} + c_t \epsilon,\quad
\forall t \geq T_\epsilon.
\end{equation*}
By deducting $\epsilon$ from both sides, we get that
\[
	\left(\delta_t - \epsilon\right) \leq \left( 1 - c_t \right)
	\left( \delta_{t-1} - \epsilon \right), \quad \forall t \geq
	T_{\epsilon}.
\]
Since $\sum c_t = \infty$, we further deduce that
\begin{align*}
\lim_{t \rightarrow \infty} \left( \delta_t - \epsilon \right)
& \leq \prod_{t = T_{\epsilon}}^{\infty} \left( 1 - c_t \right)
	\left( \delta_{T_{\epsilon} - 1} - \epsilon \right)\\
&\leq \prod_{t = T_{\epsilon}}^{\infty} \exp\left( - c_t \right)
	\left( \delta_{T_{\epsilon} - 1} - \epsilon \right)\\
& = \exp\left(-\sum_{t = T_{\epsilon}}^\infty c_t\right)
	\left( \delta_{T_{\epsilon} - 1} - \epsilon \right)
= 0,
\end{align*}
where in the first inequality we used the fact that $\exp(x) \geq 1 +
x$ for all real number $x$.
The result above then implies that
\[
	\lim_{t \rightarrow \infty} \delta_t \leq \epsilon.
\]
As $\epsilon$ is arbitrary and $\delta_t \geq 0$ from the definition,
we conclude that $\lim_{t \rightarrow \infty} \delta_t = 0$,
which is equivalent to that $W^t \rightarrow W^*$.
\end{proof}

\subsection{Proof of \cref{lemma:iterate}}
\begin{proof}
We observe that
\begin{align*}
\alpha_t^{-1} V^t &= \sum_{k=1}^t \frac{\eta_k
\beta_k}{\alpha_t } \nabla f_{\xi_k}\left(
W^{k-1} \right)\\
& = \frac{\alpha_{t-1}}{\alpha_t} \alpha_{t-1}^{-1} V^{t-1} +
	\frac{\alpha_t - \alpha_{t-1}}{\alpha_t} \nabla f_{\xi_t} \left(
	W^{t-1} \right)\\
	& = \left( 1 - \frac{\beta_t \eta_t}{\alpha_t} \right)
	\alpha_{t-1}^{-1} V^{t-1} + \frac{\beta_t
	\eta_t}{ \alpha_t}  \nabla f_{\xi_t} \left( W^{t-1} \right).
\end{align*}

From that $f$ is $L$-Lipschitz-continuously differentiable,
we have that
\begin{align}
\nonumber
\norm{ \E_{\xi_{t+1} \sim \gD} \left[ \nabla f_{\xi_{t+1}} \left(
	W^{t} \right) \right] - \E_{\xi_{t} \sim \gD} \left[ \nabla
	f_{\xi_{t}} \left( W^{t-1} \right) \right]}
&= \norm{\nabla f\left( W^t \right) - f\left( W^{t-1} \right)}\\
&\leq L \norm{W^t - W^{t-1}}.
\label{eq:changebound}
\end{align}
Therefore, \cref{eq:schedule} and \cref{eq:changebound} imply that
\begin{equation*}
	0 \leq
\frac{\norm{
	\E_{\xi_{t+1} \sim \gD} \left[ \nabla f_{\xi_{t+1}} \left(
	W^{t} \right) \right] -
	\E_{\xi_{t} \sim \gD} \left[ \nabla f_{\xi_{t}} \left(
	W^{t-1} \right) \right]}}{\beta_t \eta_t \alpha_t^{-1}}
	\leq L \frac{\norm{W^t - W^{t-1}}}{\beta_t \eta_t \alpha_t^{-1}}
	\quad \as \quad 0,
\end{equation*}
which together with the sandwich lemma shows that
\begin{equation}
\frac{\norm{
\E_{\xi_{t+1} \sim \gD} \left[ \nabla f_{\xi_{t+1}} \left(
W^{t} \right) \right] -
\E_{\xi_{t} \sim \gD} \left[ \nabla f_{\xi_{t}} \left(
W^{t-1} \right) \right]}}{\beta_t \eta_t \alpha_t^{-1}} \quad \as \quad 0.
\label{eq:convzero}
\end{equation}
Therefore, the first two conditions of \cref{eq:schedule} together
with \cref{eq:convzero} and the bounded variance assumption satisfy the requirements
of \cite[Chapter~2.4, Theorem~4.1]{AG79a}, so the conclusion of almost
sure convergence hold.

For the convergence in $\gL_2$ part, we first define $m^t \coloneqq
\alpha_t^{-1} V_t$ and $\tau_t \coloneqq \beta_t \eta_t \alpha_t^{-1}$ for notational ease.
Consider $\norm{m^{t+1}-\nabla F(W^t)}^2$, we have from the update
rule in \cref{alg:rmda} that
\begin{align}
\nonumber
&~ \norm{m^{t+1}-\nabla F(W^t)}^2 \\
\nonumber
= 
&~\norm{(1-\tau_t)m^{t}+\tau_t \nabla f_{\xi_{t+1}}(W^{t})-\nabla
F(W^{t})}^2 \\
\nonumber
= 
&~\norm{(1-\tau_t)\left(m^t- \nabla F(W^t) \right)+\tau_t \left( \nabla f_{\xi_{t+1}}(W^t) -  \nabla F(W^t) \right)}^2 \\
\nonumber
= 
&~(1-\tau_t)^2\norm{m^t- \nabla F(W^t)}^2+\tau_t^2 \norm{\nabla
	f_{\xi_{t+1}}(W^t) - \nabla  F(W^t)}^2 \\
\nonumber
&\qquad + 2 \tau_t(1-\tau_t) \inprod{m^t-\nabla F(W^t)}{\nabla
	f_{\xi_{t+1}}(W^t)-\nabla F(W^t)}\\
\label{eq:main}
= &~(1-\tau_t)^2\norm{\left(m^t- \nabla F\left( W^{t-1}
	\right)\right) + \left(\nabla F\left( W^{t-1}\right) - \nabla F \left(
	W^t \right) \right)}^2\\
	&\quad +\tau_t^2 \norm{\nabla
	f_{\xi_{t+1}}(W^t) - \nabla  F(W^t)}^2
+ 2 \tau_t(1-\tau_t) \inprod{m^t-\nabla F(W^t)}{\nabla
	f_{\xi_{t+1}}(W^t)-\nabla F(W^t)}.
\nonumber
\end{align}
Let $\{\gF_t\}_{t \geq 0}$ denote the natural
filtration of $\{(m^t,W^t)\}_{t \geq 0}$.
Namely, $\gF_t$ records the information of $W^0$,
$\{c_i\}_{i=0}^{t-1}$, $\{\eta_i\}_{i=0}^{t-1}$, and
$\{\xi_i\}_{i=1}^t$.
By defining $U_t \coloneqq \norm{m^t-\nabla F(W^{t-1})}^2$ and taking
expectation over \cref{eq:main} conditional on $\gF_t$, we obtain from
$\E\left[ \nabla f_{\xi_{t+1}}(W^t) \mid \gF_t \right] = \nabla F(W^t)
$ that
\begin{align}
\nonumber
\E \left[ U_{t+1} \mid \gF_t \right]
= &~(1-\tau_t)^2\norm{\left(m^t- \nabla F\left( W^{t-1}
	\right)\right) + \left(\nabla F\left( W^{t-1}\right) - \nabla F \left(
	W^t \right) \right)}^2 \\
&\qquad +
\tau_t^2  \E \left[ \norm{\nabla f_{\xi_{t}}(W^t) - \nabla
F(W^t)}^2 \mid \gF_t \right].
\label{eq:expmain}
\end{align}
From the last condition in \cref{eq:schedule} and the Lipschitz
continuity of $\nabla F$, there are random variables $\{ \epsilon_t
\}$ and $\{ u_t \}$ such that $\norm{u_t} = 1$, $\epsilon_t \geq 0$,
and $\nabla F(W^{t-1}) - \nabla F(W^t) = \tau_t \epsilon_t u_t$ for
all $t>0$, with $\epsilon_t \downarrow 0$ almost surely.
We thus obtain that
\begin{align}
\nonumber
&~ \norm{m^t- \nabla F(W^{t-1}) + \nabla F(W^{t-1}) - \nabla F(W^t)}^2 \\
\nonumber
=
&~ \norm{m^t- \nabla F(W^{t-1}) + \tau_t \epsilon_t u_t}^2 \\
\nonumber
=
&~(1+\tau_t)^2 \norm{ \frac{1}{1+\tau_t}\left(m^t- \nabla
F\left(W^{t-1}\right)\right) + \frac{\tau_t}{1+\tau_t} \epsilon_t u_t}^2 \\
\leq
&~ (1+\tau_t)^2 \left( \frac{1}{1+\tau_t} U_t +
\frac{\tau_t}{1+\tau_t} {\epsilon_t}^2 \right),
\label{eq:ft}
\end{align}
where we used Jensen's inequality and the convexity of
$\norm{\cdot}^2$ in the last inequality.
By substituting \cref{eq:ft} back into \cref{eq:expmain}, we obtain
\begin{align}
\nonumber
&~ \E \left[ U_{t+1} | \gF_t \right]  \\
\nonumber
\leq
&~ (1-\tau_t)^2(1+\tau_t)U_t + (1-\tau_t)^2(1+\tau_t)\tau_t
{\epsilon_t}^2 + {\tau_t}^2
\E \left[ \norm{\nabla f_{\xi_{t}}(W^t)
- \nabla  F(W^t)}^2 \mid \gF_t \right]\\
\nonumber
\leq
&~(1-\tau_t)(U_t+\tau_t{\epsilon_t}^2) + {\tau_t}^2
\E \left[ \norm{\nabla f_{\xi_{t}}(W^t)
- \nabla  F(W^t)}^2 \mid \gF_t \right]\\
\leq
&~(1-\tau_t)U_t + \tau_t {\epsilon_t}^2 + {\tau_t}^2
\E \left[ \norm{\nabla f_{\xi_{t}}(W^t)
- \nabla  F(W^t)}^2 \mid \gF_t \right].
\label{eq:lastmain}
\end{align}
For the last term in \cref{eq:lastmain}, we notice that
\begin{align}
	\nonumber
	\E \left[ \norm{\nabla f_{\xi_{t}}(W^t)
	- \nabla  F(W^t)}^2 \mid \gF_t \right] &\leq 2 \left( \E \left[
	\norm{\nabla f_{\xi_{t}}(W^t)}^2\right] + \norm{\nabla
	F(W^t)}^2\right)\\
	&\leq 2 \left(C + \norm{\nabla F(W^t)}^2 \right),
	\label{eq:intermediate}
\end{align}
where the last inequality is from the bounded variance assumption.
Since by assumption the $\{W^t\}$ lies in a bounded set $K$, we have
that for any point $W^* \in K$, $W^t - W^*$ is upper bounded,
and thus $\norm{\nabla F(W^t) - \nabla F(W^*)}$ is also bounded,
implying that $\norm{\nabla F(W^t)}^2 \leq C_2$ for some $C_2 \geq 0$.
Therefore, \cref{eq:intermediate} further leads to
\begin{equation}
	\E \left[ \norm{\nabla f_{\xi_{t}}(W^t)
	- \nabla  F(W^t)}^2 \mid \gF_t \right] \leq C_3
	\label{lm:bd}
\end{equation}
for some $C_3 \geq 0$.

Now we further take expectation on \cref{eq:lastmain} and apply
\cref{lm:bd} to obtain
\begin{align}
 \E U_{t+1}
\leq
(1-\tau_t)\E U_t + \tau_t {\epsilon_t}^2 + {\tau_t}^2 C_3
= \left( 1 - \tau_t \right) \E U_t + \tau_t \left( \epsilon_t^2 +
\tau_t C_3
\right).
\label{eq:lastmain1}
\end{align}
Note that the third implies $\epsilon_t \downarrow 0$, so this
together with the second condition that
$\tau_t \downarrow  0$
means $\epsilon_t^2 + \tau_t C_3 \downarrow 0$ as well, and
thus for any $\delta> 0$, we can find $T_\delta \geq 0$ such that
$\epsilon_t^2 + \tau_t C_3 \leq \delta$ for all $t \geq T_\delta$.
Thus, \cref{eq:lastmain1} further leads to
\begin{align}
 \E U_{t+1} - \delta
\leq
(1-\tau_t)\E U_t + \tau_t \delta - \delta
= \left( 1 - \tau_t \right) \left( \E U_t - \delta \right), \forall
t \geq T_\delta.
\label{eq:lastmain2}
\end{align}
This implies that $\left(\E U_t - \delta\right)$ becomes a decreasing
sequence starting from $t \geq T_\delta$, and since $U_t \geq 0$, this
sequence is lower bounded by $-\delta$, and hence it converges to a
certain value.
By recursion of \cref{eq:lastmain2}, we have that
\begin{align*}
 \E U_{t} - \delta
\leq
\prod_{i=T_\delta}^t (1-\tau_i)\left(\E U_{T_\delta} -
\delta\right),
\end{align*}
and from the well-known inequality $(1+x) \leq \exp^x$ for all $x \in
\gR$, the above result leads to
\begin{align*}
 \E U_{t} - \delta
\leq
\exp\left( -\sum{i=T_\delta}^t \tau_i\right) \left(\E U_{T_\delta} -
\delta\right).
\end{align*}
By letting $t$ approach infinity and noting that the first condition of
\cref{eq:schedule} indicates
\[
	\sum_{t = k}^\infty \tau_t = \infty
\]
for any $k \geq 0$, we see that
\begin{align}
-\delta \leq \lim_{t \rightarrow \infty} \E U_{t} - \delta
\leq
\exp\left( -\sum_{i=T_\delta}^\infty \tau_i\right) \left(\E U_{T_\delta} -
\delta\right)
= 0.
\label{eq:finalbound}
\end{align}
As $\delta$ is arbitrary, by taking $\delta \downarrow 0$ in
\cref{eq:finalbound} and noting the nonnegativity of $U_t$, we
conclude that $\lim \E U_t = 0$, as desired.
This proves the last result in \cref{lemma:iterate}.
\end{proof}

\subsection{Proof of \cref{thm:stationary}}
\label{app:stationary}
\begin{proof}
Using \cref{lemma:iterate}, we can view $\alpha_t^{-1} V^t$ as
$\nabla f(W^t)$ plus some noise that asymptotically decreases to zero
with probability one:
\begin{equation}
	\alpha_t^{-1} V_t = \nabla f(W^t) + \epsilon_t,\quad
	\norm{\epsilon_t} \xrightarrow{\text{a.s.}} 0.
	\label{eq:grad}
\end{equation}
We use \cref{eq:grad} to rewrite the optimality condition of
\cref{eq:daprox} as (also see \cref{eq:prox} of \cref{alg:rmda})
\begin{equation}
	-\left(\nabla f\left( W^t \right) + \epsilon_t + \beta_t \alpha_t^{-1} \left( \tilde
	W^{t} - W^0 \right) \right) \in \partial \psi \left( \tilde
	W^{t}
	\right).
	\label{eq:opt}
\end{equation}
Now we consider $\partial F(\tilde W^t)$.
Clearly from \cref{eq:opt}, we have that
\begin{equation}
	\nabla f\left( \tilde W^t \right) - \nabla f\left( W^t \right) - \epsilon_t - \beta_t \alpha_t^{-1} \left( \tilde
	W^{t} - W^0 \right)\in \partial \nabla f\left( \tilde W^t
	\right) + \psi \left( \tilde W^{t}\right) = \partial F\left(
	\tilde W^t \right) .
	\label{eq:opt2}
\end{equation}
Now we consider the said event that $\tilde W^t \rightarrow W^*$ for a
certain $W^*$, and let us define this event as $\gA \subseteq \Omega$.
From \cref{lemma:conv}, we know that $W^t \rightarrow W^*$ as well
under $\gA$.
Let us define $\gB \subseteq \Omega$ as the event of $\epsilon_t \rightarrow 0$,
then we know that since $P(\gA) > 0$ and $P(\gB) = 1$, where $P$ is
the probability function for events in $\Omega$, $P(\gA \cap
\gB) = P(\gA)$.
Therefore, conditional on the event of $\gA$, we have that $\epsilon_t
\as 0$ still holds.
Now we consider any realization of $\gA \cap \gB$.
For the right-hand side of \cref{eq:opt2},
as $\tilde W^{t}$ is convergent and $\beta_t \alpha_t^{-1}$ decreases
to zero, by letting $t$ approach infinity,
we have that
\[
	\lim_{t \rightarrow \infty} \epsilon_t + \beta_t \alpha_t^{-1}
	\left( \tilde W^t - W^0 \right) = 0 + 0 \left( W^* - W^0 \right) =
	0.
\]
By the Lipschitz continuity of $\nabla f$, we have
from \cref{eq:update} and \cref{eq:schedule} that
\[
	0 \leq \norm{\nabla f\left( \tilde W^t \right) - \nabla f\left( W^t
	\right)}
	\leq L \norm{ W^t -  \tilde W^t}.
\]
As $\{W^t\}$ and $\{\tilde W^t\}$ converge to the same point,
we see that $\norm{W^t - \tilde W^t} \rightarrow 0$, so $\nabla
f\left( \tilde W^t \right) - \nabla f(W^t)$ also approaches zero.
Hence, the limit of the right-hand side of \cref{eq:opt2} is
\begin{equation}
	\lim_{t \rightarrow \infty}
	\nabla f\left( \tilde W^t \right)-\left(\nabla f\left( W^t \right) + \epsilon_t + \beta_t \alpha_t^{-1} \left( \tilde
	W^{t} - W^0 \right) \right)
	= 0
	\label{eq:rhs}.
\end{equation}
On the other hand, for the left-hand side of \cref{eq:opt2}, the outer
semicontinuity of $\partial \psi$ at $W^*$ and the continuity of
$\nabla f$ show that
\begin{equation}
	\lim_{t \rightarrow \infty} \nabla f(\tilde W^t) + \partial \psi(\tilde W^t) \subseteq
	\partial \nabla f(W^*) + \psi \left( W^* \right) = \partial F(W^*).
	\label{eq:lhs}
\end{equation}
Substituting \cref{eq:rhs} and \cref{eq:lhs} back into \cref{eq:opt2}
then proves that $0 \in \partial F(W^*)$ and thus $W^* \in
\gZ$.
\end{proof}

\subsection{Proof of \cref{thm:identify}}
\begin{proof}
Our discussion in this proof are all under the event that $\tilde W^t \rightarrow W^*$.
From the argument in \cref{app:stationary}, we can view $\alpha_t^{-1}
V^t$ as $\nabla f(W^t)$ plus some noise that asymptotically decreases
to zero with probability one as shown in \cref{eq:grad}.
From \cref{lemma:conv}, we know that $W^t \rightarrow W^*$.
From \cref{eq:opt}, there is $U^t \in \partial \psi\left( \tilde W^t
\right)$ such that
\begin{equation}
U^t = - \alpha_t^{-1} V^t + \alpha_t^{-1} \beta_t \left( \tilde W^t -
	W^0 \right).
\label{eq:partial}
\end{equation}
Moreover, we define
\begin{equation}
	\gamma_t \coloneqq W^t - \tilde W^t.
	\label{eq:noise}
\end{equation}
By combining \crefrange{eq:partial}{eq:noise} with \cref{eq:grad},
we obtain
\begin{align}
\nonumber
&~\min_{Y \in \partial F(\tilde W^t)} \norm{Y}\\
\nonumber
\leq &~\norm{\nabla f \left( \tilde W^t \right) + U^t}\\
\nonumber
= &~\norm{\nabla f \left( \tilde W^t \right) - \nabla f\left( W^t
	\right) - \epsilon_t - \alpha_t^{-1} \beta_t \left( \tilde W^t -
	W^0 \right)}\\
\nonumber
\leq & \norm{\nabla f \left( \tilde W^t \right) - \nabla f\left( W^t
	\right)} + \norm{\epsilon_t} + \alpha_t^{-1} \beta_t \norm{\tilde
	W^t - W^0}\\
\leq & L \norm{\gamma_t} + \norm{\epsilon_t} + \alpha_t^{-1} \beta_t
\left(\norm{W^* - \tilde W^t} + \norm{W^0 - W^*}\right),
\label{eq:tobound}
\end{align}
where we used the Lipschitz continuity of $\nabla f$ and the triangle
inequality in the last inequality.

We now separately bound the terms in \cref{eq:tobound}.
From that $W^t \rightarrow W^*$ and $\tilde W^t \rightarrow W^*$, it is
straightforward that $\norm{\gamma_t} \rightarrow 0$.
The second term decreases to zero almost surely according to
\cref{eq:grad} and the argument in \cref{app:stationary}.
For the last term, since $\alpha_t^{-1} \beta_t \rightarrow 0$, and
$\norm{\tilde W^t - W^*} \rightarrow 0$, we know that
\[
	\alpha_t^{-1} \beta_t \norm{W^0 - W^*} \rightarrow 0,\quad
	\alpha_t^{-1} \beta_t \norm{\tilde W^t - W^*} \rightarrow 0.
\]
Therefore, we conclude from the above argument and \cref{eq:tobound}
that
\begin{equation*}
\min_{Y \in \partial F(\tilde W^t)} \norm{Y} \as 0.
\end{equation*}
As $f$ is smooth with probability one,
we know that if $\psi$ is partly smooth at $W^*$ relative to $\gM$,
then so is $F = f + \psi$ with probability one.
Moreover, Lipschitz-continuously differentiable functions are always
prox-regular, and the sum of two prox-regular functions is still
prox-regular, so $F$ is also prox-regular at $W^*$ with probability one.
Following the argument identical to that in \cref{app:stationary},
we know that these probability one events are still probability one
conditional on the event of $\tilde W^t \rightarrow W^*$ as this event
has a nonzero probability.
As $\tilde W^t \rightarrow W^*$ and $\nabla f(\tilde W^t) + U^t \as 0 \in
\partial F(W^*)$ (the inclusion is from \cref{eq:nod}), we have from
the subdifferential continuity of $\psi$ and the smoothness of $f$
that $F(\tilde W^t) \as F(W^*)$.  Since we also have $\tilde W^t
\rightarrow W^*$ and $\min_{Y \in \partial F(\tilde W^t)} \norm{Y} \as
0$,
clearly
\begin{equation}
\left(\tilde W^t, F \left( W^t \right), \min_{Y \in \partial F\left(
\tilde W^t \right)}
\norm{Y}\right) \as \left(W^*, F(W^*), 0\right).
\label{eq:condas}
\end{equation}
Therefore, \cref{eq:condas} and \cref{eq:nod} together with the
assumptions on $\psi$ at $W^*$ imply that with probability one, all
conditions of Lemma~1 of \cite{CPL20b} are satisfied, so from it,
\cref{eq:identify} holds almost surely, conditional on the event of
$\tilde W^t \rightarrow W^*$.
\end{proof}

\section{Additional Discussions on Applications}
\label{app:applications}
We now discuss in more technical details the applications in
\cref{sec:sparse}, especially regarding how the regularizers satisfy
the properties required by our theory.

\subsection{Structured Sparsity}
\label{app:sparse}
\begin{figure}[tb]
	\centering
	\includegraphics[width=3in]{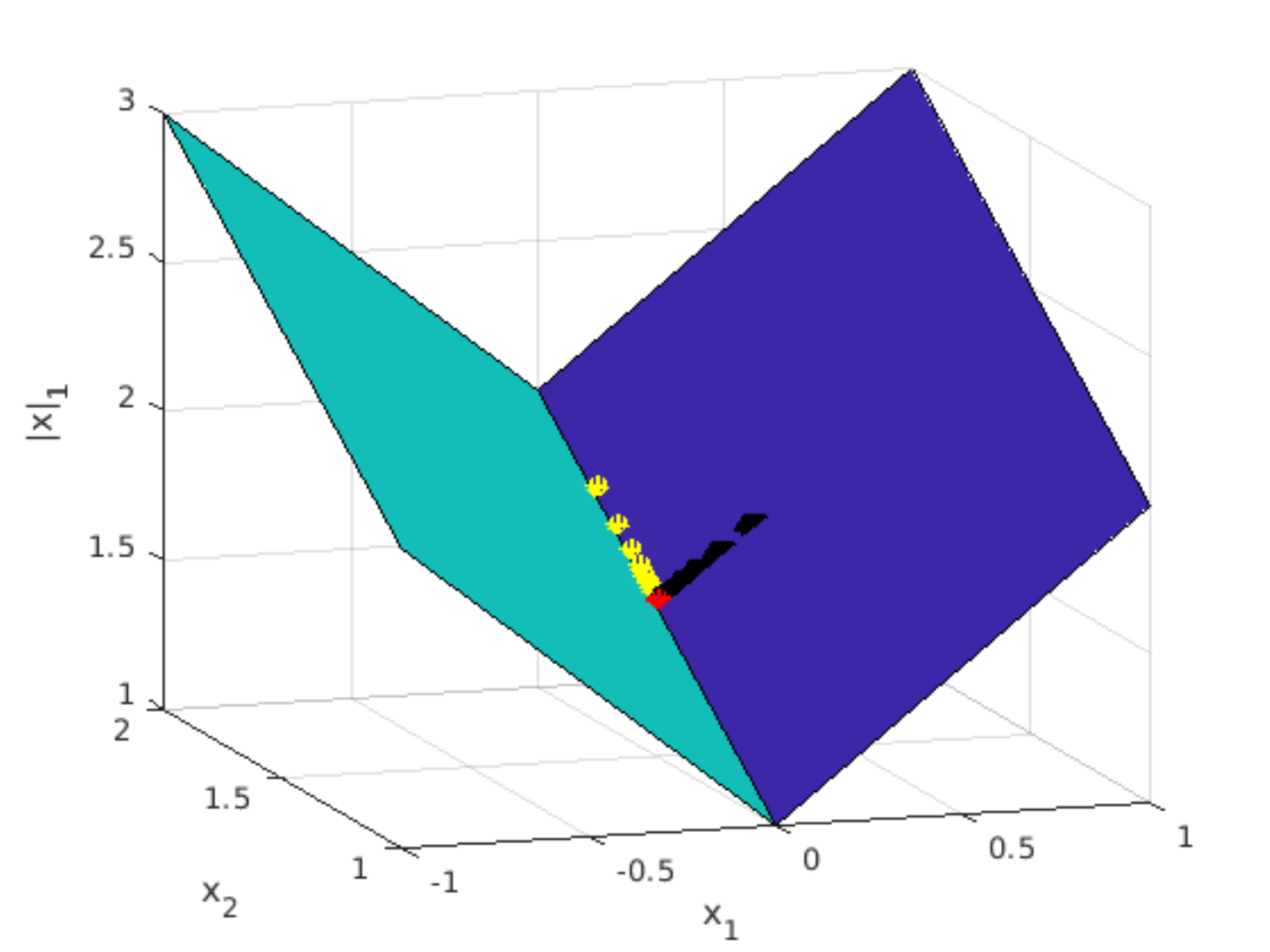}
	\caption{An illustration of partial smoothness of the $\ell_1$ norm.}
	\label{fig:l1}
\end{figure}

We start our discussion with the simple $\ell_1$ norm as the warm-up
for the group-LASSO norm.
It is clear that $\norm{W}_1$ is a convex function that is finite
everywhere, so it is prox-regular, subdifferentially continuous, and
regular everywhere, hence we just need to discuss about the remaining
parts in \cref{def:ps}.
Consider a problem with dimension $n > 0$.
Note that
\[
	\norm{x}_1 = \sum_{i=1}^n |x_i|,
\]
and the absolute value is smooth everywhere except the point of
origin.
Therefore, it is clear that $\norm{x}_1$ is locally smooth if $x_i
\neq 0$ for all $i$.
For any point $x^*$,
when there is an index set $I$ such that $x^*_i = 0$ for all $i \in I$
and $x^*_i \neq 0$ for $i \notin I$, we see that the part of the norm
corresponds to $I^C$ (the complement of $I$):
\[
	\sum_{i \in I^C} |x^*_i|
\]
is locally smooth around $x^*$.
Without loss of generality, we assume that $I = \{1,2,\dotsc,k\}$ for
some $k \geq 0$, then the subdifferential of $\norm{x}_1$ at $x^*$ is
the set
\begin{equation}
	\{\text{sgn}(x_1)\}\times \cdots \times \{\text{sgn}(x_k)\} \times
	[-1,1]^{n - k},
	\label{eq:subd}
\end{equation}
and clearly if we move from $x^*$ along any direction
$y \coloneqq (y_1,\dotsc,y_k,0,\dotsc,0)$ with a small step,
the function value changes smoothly as it is a linear function,
satisfying the first condition of \cref{def:ps}.
Along the same direction $y$ with a small enough step, the set of
subdifferential remains the same, so the continuity of subdifferential
requirement holds.
We can also observe from the above argument that the manifold should
be $\mathcal{M}_{x^*} = \{x \mid x_i = 0, \forall i \in I\}$, and
clearly it is a subspace of $\R^n$ with its normal space at $x^*$ being
$N \coloneqq \{y \mid \inprod{x^*}{y} = 0\} = \{y \mid y_i = 0,
\forall i \in I^C\}$, which is clearly the affine span of
\cref{eq:subd} with the translation being $(\text{sgn}(x_1)\times
\cdots \times \text{sgn}(x_k), 0,\dotsc,0)$.
Moreover, indeed the manifolds are low dimensional ones, and for
iterates approaching $x^*$, staying in this active manifold means that
the (unstructured) sparsity of the iterates is the same as the limit
point $x^*$.
We also provide a graphical illustration of $\norm{x}_1$ with $n=2$ in
\cref{fig:l1}.
We can observe that for any $x$ with $x_1 \neq 0$ and $x_2 \neq 0$,
the function is smooth locally around any point, meaning that
$\norm{x}_1$ is partly smooth relative to the whole space at $x$ (so
actually smooth locally around $x$).
For $x$ with $x_1 = 0$, the function value corresponds to the sharp
valley in the graph, and we can see that the function is smooth along
the valley, and this valley corresponds to the one-dimensional
manifold $\{x \mid x_1 = 0\}$ for partial smoothness.

Next, we use the same graph to illustrate the importance of manifold
identification.
Consider that the red point $x^* = (0,1.5)$ is the limit point of the
iterates of a certain algorithm, and the yellow points and black
points are two sequences that both converge to $x^*$.
If the iterates of the algorithm are the black points, then clearly
except for the limit point itself, all iterates are nonsparse,
and thus the final output of the algorithm is also nonsparse unless we
can get to exactly the limit point within finite iterations (which is
usually impossible for iterative methods).
On the other hand, if the iterates are the yellow points, this is the
case that the manifold is identified, because all points sit in the
valley and enjoy the same sparsity pattern as the limit point $x^*$.
This is why we concern about manifold identification when we solve
regularized optimization problems.

From this example, we can also see an explanation for why our
algorithm with the property of manifold identification performs better
than other methods without such a property.
Consider a Euclidean space any point $x^*$ with an index set $I$ such
that $x^*_I = 0$ and $|I| > 0$.
This means that $x^*$ has at least one coordinate being zero, namely
$x^*$ contains sparsity.
Now let
\[
	\epsilon_0 \coloneqq \min_{i \in I^C}\quad |x^*_i|,
\]
then from the definition of $I$, $\epsilon_0 > 0$.
Fro any sequence $\{x^t\}$ converging to $x^*$,
for any $\epsilon \in (0, \epsilon_0)$, we can find $T_\epsilon \geq
0$ such that
\[
	\norm{x^t-x^*}_2 \leq \epsilon, \quad \forall t \geq T_\epsilon.
\]
Therefore, for any $i \notin I$, we must have that $x^t_i \neq 0$ for
all $t \geq T_\epsilon$.
Otherwise, $\norm{x^t - x^*}_2 \geq \epsilon_0$, but $\epsilon_0 >
\epsilon \geq \norm{x^t-x^*}_2$, leading to a contradiction.
On the other hand, for any $i \in I$, we can have $x^t_i \neq 0$ for
all $t$ without violating the convergence.
That being said, for any sequence converging to $x^*$, eventually the
iterates cannot be sparser than $x^*$, so the sparsity level of $x^*$,
or of its active manifold, is the local upper bound for the sparsity
level of points converging to $x^*$.
Therefore, if iterates of two algorithms converge to the same limit
point, the one with a proven manifold identification ability clearly
will produce a higher sparsity level.

Similar to our example here, in applications other than sparsity,
iterates converging to a limit point dwell on super-manifolds of the
active manifold, and the active manifold is the minimum one that locally
describes points with the same structure as the limit point, and thus
identifying this manifold is equivalent to finding the locally most ideal
structure of the application.

Now back to the sparsity case.
One possible concern is the case that the limit point is $(0,0)$ in
the two-dimension example.
In this case, the manifold is the $0$-dimensional subspace $\{0\}$.
If this is the case and manifold identification can be ensured, it
means that limit point itself can be found within finite iterations.
This case is known as the weak sharp minima \citep{BurF93a} in
nonlinear optimization, and its associated finite termination property
is also well-studied.

For this example, We also see that $\norm{x}_1$ is partly smooth at
any point $x^*$, but the manifold differs with $x^*$.
This is a specific benign example, and in other cases, partial
smoothness might happen only locally at some points of interest
instead of everywhere.

Next, we further extend our argument above to the case of
\cref{eq:grouplasso}.
This can be viewed as the $\ell_1$ norm for each group and we can
easily obtain similar results.
Again, since the group-LASSO norm is also convex and finite
everywhere, prox-regularity, regularity, and subdifferential
continuity are not issues at all.
For the other properties, we consider one group first, then the
group-LASSO norm reduces to the $\ell_2$ norm.
Clearly, $\norm{x}_2$ is smooth locally if $x \neq 0$, with the
gradient being $x/\norm{x}_2$, but it is nonsmooth at the point $x =
0$, where the subdifferential is the unit ball.
This is very similar to the absolute value, whose subdifferential at
$0$ is the interval $[-1,1]$.
Thus, we can directly apply similar arguments above, and conclude that
for any $W^*$, \cref{eq:grouplasso} is partly smooth at $W^*$ with respect
to the manifold
$\mathcal{M}_{W^*} = \{W \mid W_{\gI_g} = 0, \forall g: W^*_{\gI_g} =
0\}$,
which is again a lower-dimensional subspace.
Therefore, the manifold of defining the partial smoothness for the
group-LASSO norm exactly corresponds to its structured sparsity pattern.

\subsection{Binary Neural Networks}
We continue to consider the binary neural network problem.
For easier description, for the Euclidean space $\gE$ we consider, we
will use a vectorized representation for $W, A \in \gE$ such that
the elements are enumerated as $W_1,\dotsc,W_n$ and
$\alpha_1,\dotsc,\alpha_n$.
The corresponding optimization problem can therefore be written as
\begin{equation}
\min_{W,A \in \gE}\quad \E_{\xi\sim \gD}
\left[ f_{\xi} \left( W \right) \right] + \lambda \sum_{i=1}^n
\left(\alpha_i\left(w_i + 1\right)^2 + \left(1 -
\alpha_i\right)\left(w_i - 1\right)^2+ \delta_{[0,1]}\left( \alpha_i
\right)\right),
\label{eq:binary}
\end{equation}
where given any set $C$, $\delta_C$ is the indicator function of $C$,
defined as
\[
	\delta_{C}(x) = \begin{cases}
		0 & \text{ if } x \in C,\\
		\infty & \text{ else}.
	\end{cases}
\]
We see that except for the indicator function part, the objective is
smooth, so the real partly smooth term that we treat as the
regularizer is
\[
	\Phi(\alpha) \coloneqq \sum_{i=1}^n \delta_{[0,1]}(\alpha_i).
\]
We note that for $\alpha_i \in (0,1)$, the value of
$\delta_{[0,1]}(\alpha_i)$ remains a constant zero in a neighborhood
of $\alpha_i$, and for $\alpha_i
\notin [0,1]$, the indicator function is also constantly infinite
within a neighborhood.
Thus, the point of nonsmoothness, happens only at $\alpha_i \in
\{0,1\}$, and similar to our discussion in the previous subsection,
$\Phi$ is partly smooth along directions that we fix those $\alpha_i$
at the boundary (namely, being either $0$ or $1$) unchanged.
The identified manifold therefore corresponds to the entries of
$\alpha_i$ that are fixed at $0$ or $1$, and this can serve as the
indicator for the desired binary pattern in this task.

\section{Experiment Setting Details}
\label{sec:expdetails}
For the weights $w_g$ of each group in \cref{eq:grouplasso}, for all
experiments in \cref{sec:exp}, we follow \cite{deleu2021structured} to
set $w_g = \sqrt{|\gI_g|}$.
All \ssi parameter settings, excluding the regularization weight and the
learning rate schedule, follow the default values in their package.

\crefrange{tbl:Lin_MNIST}{tbl:VGG19} provide detailed
settings of \cref{sec:NN}.
For the modified VGG19 model, we follow \citet{deleu2021structured} to
eliminate all fully-connected layers except the output layer, and add
one batch-norm layer \citep{ioffe2015batch} after each convolutional
layer to simulate modern CNNs like those proposed in \cite{he2016deep,
huang2017densely}.
For ResNet50 in the structured sparsity experiment in \cref{sec:NN},
our version of ResNet50 is the one constructed by the publicly
available script at
\url{https://github.com/weiaicunzai/pytorch-cifar100}.

In the unstructured sparsity experiment presented in \cref{sec:cp},
for better comparison with existing works in the literature of
pruning,
we adopt the version of ResNet50 used by
\cite{VS21a}.\footnote{\url{https://github.com/varun19299/rigl-reproducibility.}}
\cref{tbl:UNSS} provides detailed settings of \cref{sec:cp}.
For \rigl, we use the PyTorch implementation of \citet{VS21a}.

\begin{table}[tbh]
\centering
\caption{Details of the experimental settings of logistic regression
	on MNIST in \cref{sec:NN}.}
\label{tbl:Lin_MNIST}
\begin{tabular}{|l|l|}
	\hline  
	Parameter & Value \\
	\hline  
	Data set & MNIST \\
	Model & Logistic regression \\
	Loss function & Cross entropy \\
	Regularization function & Group LASSO \\
	Regularization weight & $10^{-3}$ \\
	Total epochs & 500 \\
	\hline
	\hline
	\multicolumn{2}{|c|}{\sgd} \\
	\hline
	Learning rate schedule & $\eta(\text{epoch}) =
	10^{-1- \lfloor\text{epoch} / 50\rfloor}$\\
	Momentum &  $10^{-1}$\\
	\hline
	\multicolumn{2}{|c|}{\ssi} \\
	\hline
	Learning rate schedule & $\eta(\text{epoch}) = 10^{-3-
		\lfloor\text{epoch} / 50\rfloor}$\\
	\hline
	\multicolumn{2}{|c|}{\rmda} \\
	\hline
	Restart epochs & $50, 100, 150, 200$\\
	Learning rate schedule & $\eta(\text{epoch}) = \max(10^{-5},10^{-1-
		\lfloor\text{epoch} / 50\rfloor})$\\
	Momentum schedule & $c(\text{epoch}) = \min(1, 10^{-2 +
		\lfloor\text{epoch} / 50\rfloor })$\\
	\hline	
\end{tabular}
\end{table}

\begin{table}[tbh]
\centering
\caption{Details of the experimental settings of the multi-layer
	fully-connected NN on FashionMNIST in \cref{sec:NN}.}
\label{tbl:MFL_FashionMNIST}
\begin{tabular}{|l|l|}
	\hline  
	Parameter & Value \\\hline
	Data set & FashionMNIST \\
	Model & Fully-connected NN (\cref{tbl:MFL}) \\
	Loss function & Cross entropy \\
	Regularization function & Group LASSO \\
	Total epochs & 500 \\
	\hline  
	\hline
	\multicolumn{2}{|c|}{\sgd} \\
	\hline
	Regularization weight & $10^{-4}$ \\
	Learning rate schedule & $\eta(\text{epoch}) = 10^{-1 - \lfloor	\text{epoch} / 50 \rfloor}$ \\
	Momentum & $10^{-1}$ \\
	\hline
	\multicolumn{2}{|c|}{\ssi} \\
	\hline
	Regularization weight & $4 \times 10^{-6}$ \\
	Learning rate schedule & $\eta(\text{epoch}) = 10^{-3- \lfloor
		\text{epoch} / 50 \rfloor}$\\
	\hline
	\multicolumn{2}{|c|}{\rmda} \\
	\hline
	Regularization weight & $7 \times 10^{-5}$ \\
     Restart epochs & $50, 100, 150, 200$\\
	Learning rate schedule & $\eta(\text{epoch}) = \max(10^{-5},
	10^{-1 -\lfloor		\text{epoch} / 50 \rfloor})$\\
     Momentum schedule & $c(\text{epoch}) = \min(1,10^{-2 + \lfloor
		 \text{epoch} / 50 \rfloor})$\\
	\hline	
\end{tabular}
\end{table}

\begin{table}[tbh]
\centering
\caption{Details of the experimental settings of LeNet5 on MNIST in \cref{sec:NN}}
\label{tbl:LeNet5_MNIST}
\begin{tabular}{|@{}l|l|}
	\hline  
	Parameter & Value \\\hline
	Data set & MNIST \\
	Model & LeNet5 (\cref{tbl:LeNet5_large}) \\
	Loss function & Cross entropy \\
	Regularization function & Group LASSO \\
	Total epochs & 500\\
	\hline  
	\hline
	\multicolumn{2}{|c|}{\sgd} \\
	\hline
	Regularization weight & $1.2 \times 10^{-4}$ \\
	Learning rate schedule & $\eta(\text{epoch}) =
	10^{-1 - \lfloor\text{epoch} / 50\rfloor}$ \\
	Momentum & $10^{-1}$  \\
	\hline
	\multicolumn{2}{|c|}{\ssi} \\
	\hline
	Regularization weight & $9 \times 10^{-5}$ \\
	Learning rate schedule &
	$\eta(\text{epoch}) = 10^{-3 - \lfloor\text{epoch} / 50\rfloor}$
	\\
	\hline
	\multicolumn{2}{|c|}{\rmda} \\
	\hline
	Regularization weight & $10^{-4}$\\
     Restart epochs & $50, 100, 150, 200$\\
	Learning rate schedule &
	$\eta(\text{epoch}) = \max(10^{-4},10^{- \lfloor\text{epoch} /
50\rfloor})$\\
     Momentum schedule &
	$c(\text{epoch}) = \min(1,10^{-2 + \lfloor\text{epoch} /
50\rfloor})$\\
	\hline	
\end{tabular}
\end{table}

\begin{table}[tbh]
\centering
\caption{Details of the experimental settings of LeNet5 on FashionMNIST in \cref{sec:NN}}
\label{tbl:LeNet5_FashionMNIST}
\begin{tabular}{|@{}l|l|}
	\hline  
	Parameter & Value \\\hline
	Data set & FashionMNIST \\
	Model & LeNet5 (\cref{tbl:LeNet5_large}) \\
	Loss function & Cross entropy \\
	Regularization function & Group LASSO \\
	Total epochs & 500\\
	\hline  
	\hline
	\multicolumn{2}{|c|}{\sgd} \\
	\hline
	Regularization weight & $1.2 \times 10^{-4}$ \\
	Learning rate schedule & $\eta(\text{epoch}) =
	10^{-1 - \lfloor\text{epoch} / 50\rfloor}$ \\
	Momentum & $10^{-1}$  \\
	\hline
	\multicolumn{2}{|c|}{\ssi} \\
	\hline
	Regularization weight & $6 \times 10^{-5}$ \\
	Learning rate schedule & $\eta(\text{epoch}) = 10^{-3 -
		\lfloor\text{epoch} / 50\rfloor}$ \\
	\hline
	\multicolumn{2}{|c|}{\rmda} \\
	\hline
	Regularization weight & $10^{-4}$\\
     Restart epochs & $50, 100, 150, 200$\\
	Learning rate schedule &
	$\eta(\text{epoch}) = \max(10^{-4},10^{- \lfloor\text{epoch} /
50\rfloor})$\\
     Momentum schedule &
	$c(\text{epoch}) = \min(1,10^{-2 + \lfloor\text{epoch} /
50\rfloor})$\\
	\hline	
\end{tabular}
\end{table}

\begin{table}[tbh]
\centering
\caption{Details of the experimental settings of the modified VGG19 on
	CIFAR10 in \cref{sec:NN}.}
\label{tbl:VGG19_CIFAR10}
\begin{tabular}{|@{}l|l|}
	\hline  
	Parameter & Value \\\hline
	Data set & CIFAR10 \\
	Model & VGG19 (\cref{tbl:VGG19}) \\
	Loss function & Cross entropy \\
	Regularization function & Group LASSO \\
	Total epochs & 1000 \\
	\hline  
	\hline
	\multicolumn{2}{|c|}{\sgd} \\
	\hline
	Regularization weight & $5 \times 10^{-5}$ \\
	Learning rate schedule & $\eta(\text{epoch}) =
	10^{-1 - \lfloor\text{epoch} / 100\rfloor}$ \\
	Momentum & $10^{-1}$  \\
	\hline
	\multicolumn{2}{|c|}{\ssi} \\
	\hline
	Regularization weight & $3 \times 10^{-7}$ \\
	Learning rate schedule & $\eta(\text{epoch}) =
	10^{-3 - \lfloor\text{epoch} / 100\rfloor}$ \\
	\hline
	\multicolumn{2}{|c|}{\rmda} \\
	\hline
	Regularization weight& $10^{-4}$ \\
     Restart epochs & $100, 200, 300, 400, 500$\\
	Learning rate schedule &
	$\eta(\text{epoch}) = \max(10^{-6},10^{-1 - \lfloor\text{epoch} /
100\rfloor})$\\
     Momentum schedule &
	$c(\text{epoch}) = \min(1,10^{-2 + \lfloor\text{epoch} /
100\rfloor})$\\
	\hline	
\end{tabular}
\end{table}

\begin{table}[tbh]
\centering
\caption{Details of the experimental settings of the modified VGG19 on
	CIFAR100 in \cref{sec:NN}.}
\label{tbl:VGG19_CIFAR100}
\begin{tabular}{|@{}l|l|}
	\hline  
	Parameter & Value \\\hline
	Data set & CIFAR100 \\
	Model & VGG19 (\cref{tbl:VGG19}) \\
	Loss function & Cross entropy \\
	Regularization function & Group LASSO \\
	Total epochs & 1000 \\
	\hline  
	\hline
	\multicolumn{2}{|c|}{\sgd} \\
	\hline
	Regularization weight & $3 \times 10^{-5}$ \\
	Learning rate schedule & $\eta(\text{epoch}) =
	10^{-1 - \lfloor\text{epoch} / 100\rfloor}$ \\
	Momentum & $10^{-1}$  \\
	\hline
	\multicolumn{2}{|c|}{\ssi} \\
	\hline
	Regularization weight & $10^{-7}$ \\
	Learning rate schedule & $\eta(\text{epoch}) =
	10^{-3 - \lfloor\text{epoch} / 100\rfloor}$ \\
	\hline
	\multicolumn{2}{|c|}{\rmda} \\
	\hline
	Regularization weight& $10^{-4}$ \\
	Restart epochs & $100, 200, 300, 400, 500$\\
	Learning rate schedule &
	$\eta(\text{epoch}) = \max(10^{-6},10^{-1 - \lfloor\text{epoch} /
100\rfloor})$\\
Momentum schedule &
	$c(\text{epoch}) = \min(1,10^{-2 + \lfloor\text{epoch} /
100\rfloor})$\\
	\hline	
\end{tabular}
\end{table}

\begin{table}[tbh]
\centering
\caption{Details of the experimental settings of ResNet50 on CIFAR10 in \cref{sec:NN}.
We use the ResNet50 model from the public script
\url{https://github.com/weiaicunzai/pytorch-cifar100}.}
\label{tbl:ResNet50_CIFAR10}
\begin{tabular}{|@{}l|l|}
	\hline  
	Parameter & Value \\\hline
	Data set & CIFAR10 \\
	Model & ResNet50 \\
	Loss function & Cross entropy \\
	Regularization function & Group LASSO \\
	Total epochs & 1500 \\
	\hline  
	\hline
	\multicolumn{2}{|c|}{\sgd} \\
	\hline
	Regularization weight & $5 \times 10^{-5}$ \\
	Learning rate schedule & $\eta(\text{epoch}) =
	10^{-1 - \lfloor\text{epoch} / 150\rfloor}$ \\
	Momentum & $10^{-1}$  \\
	\hline
	\multicolumn{2}{|c|}{\ssi} \\
	\hline
	Regularization weight & $3 \times 10^{-7}$ \\
	Learning rate schedule & $\eta(\text{epoch}) =
	10^{-3 - \lfloor\text{epoch} / 150\rfloor}$ \\
	\hline
	\multicolumn{2}{|c|}{\rmda} \\
	\hline
	Regularization weight& $10^{-5}$ \\
	Restart epochs & $150, 300, 450, 600$\\
	Learning rate schedule &
	$\eta(\text{epoch}) = \max(10^{-4},10^{-\lfloor\text{epoch} /
150\rfloor})$\\
Momentum schedule &
	$c(\text{epoch}) = \min(1,10^{-2 + \lfloor\text{epoch} /
150\rfloor})$\\
	\hline	
\end{tabular}
\end{table}

\begin{table}[tbh]
\centering
\caption{Details of the experimental settings of ResNet50 on CIFAR100 in \cref{sec:NN}.
We use the ResNet50 model from the public script
\url{https://github.com/weiaicunzai/pytorch-cifar100}.}
\label{tbl:ResNet50_CIFAR100}
\begin{tabular}{|@{}l|l|}
	\hline  
	Parameter & Value \\\hline
	Data set & CIFAR100 \\
	Model & ResNet50 \\
	Loss function & Cross entropy \\
	Regularization function & Group LASSO \\
	Total epochs & 1500 \\
	\hline  
	\hline
	\multicolumn{2}{|c|}{\sgd} \\
	\hline
	Regularization weight & $4 \times 10^{-5}$ \\
	Learning rate schedule & $\eta(\text{epoch}) =
	10^{-1 - \lfloor\text{epoch} / 150\rfloor}$ \\
	Momentum & $10^{-1}$  \\
	\hline
	\multicolumn{2}{|c|}{\ssi} \\
	\hline
	Regularization weight & $3 \times 10^{-7}$ \\
	Learning rate schedule & $\eta(\text{epoch}) =
	10^{-3 - \lfloor\text{epoch} / 150\rfloor}$ \\
	\hline
	\multicolumn{2}{|c|}{\rmda} \\
	\hline
	Regularization weight& $10^{-5}$ \\
     Restart epochs & $150, 300, 450, 600$\\
	Learning rate schedule &
	$\eta(\text{epoch}) = \max(10^{-4},10^{-\lfloor\text{epoch} /
150\rfloor})$\\
     Momentum schedule &
	$c(\text{epoch}) = \min(1,10^{-2 + \lfloor\text{epoch} /
150\rfloor})$\\
	\hline	
\end{tabular}
\end{table}

\begin{table}[tbh]
\centering
\caption{Details of the multi-layer fully-connected NN.
\url{https://github.com/zihsyuan1214/rmda/blob/master/Experiments/Models/mlp.py}.
}
\label{tbl:MFL}
\begin{tabular}{|l|l|}
	\hline  
	Parameter & Value \\\hline
	Type of layers & fully-connected layer \\
	Number of layers & 7 \\
	Number of output neurons each layer: $1,2,3,4,5,6,7$ &
	$512,256,128,64,32,16,10$ \\
	Activation function for convolution/output layer & relu/softmax \\
	\hline
	\end{tabular}
\end{table}

\begin{table}[tbh]
\centering
\caption{Details of the modified LeNet5 for experiments in
	\cref{sec:NN}.
\url{https://github.com/zihsyuan1214/rmda/blob/master/Experiments/Models/lenet5_large.py}.}
\label{tbl:LeNet5_large}
\begin{tabular}{|l|l|}
	\hline  
	Parameter & Value \\\hline
	Number of layers & 4 \\
	Number of convolutional layers &  2\\
	Number of fully-connected layers & 2 \\
	Size of convolutional kernels & $3 \times 3$ \\
	Number of output filters $1, 2$ & $20, 50$ \\
	Number of output neurons $3, 4$ & $500, 10$ \\
	Kernel size, stride, padding of maxing pooling & $2 \times 2$, none, invalid\\
	Operations after convolutional layers & max pooling \\
	Activation function for convolution/output layer & relu/softmax \\
	\hline
\end{tabular}
\end{table}

\begin{table}[tbh]
\centering
\caption{Details of the modified VGG19.
\url{https://github.com/zihsyuan1214/rmda/blob/master/Experiments/Models/vgg19.py}.}
\label{tbl:VGG19}
\begin{tabular}{|l|l|}
	\hline  
	Parameter & Value \\\hline
	Number of layers & 17 \\
	Number of convolutional layers & 16 \\
	Number of fully-connected layers & 1 \\
	Size of convolutional kernels & $3\times3$ \\
	Number of output filters 1-2, 3-4, 5-8, 9-16 & $64, 128, 256,
	512$ \\
	Kernel size, stride, padding of maxing pooling & $2\times 2, 2$, invalid\\
	Operations after convolutional layers & max pooling, batchnorm \\
	Activation function for convolution/output layer & relu/softmax \\
	\hline
\end{tabular}
\end{table}

\begin{table}[tbh]
\centering
\caption{Details of experimental settings of ResNet50 on CIFAR10 and
	CIFAR100 for unstructured sparsity.
In this experiment, we adopt the version of ResNet50 in \cite{VS21a}.}
\label{tbl:UNSS}
\begin{tabular}{|l|l|}
	\hline  
	Parameter & Value \\\hline
	Model & ResNet50 \\
	Loss function & Cross entropy \\
	\hline
	\multicolumn{2}{|c|}{\rmda} \\
	\hline
	Data & CIFAR10 \\
	Total epochs & 1000 \\
	L1 weight & $10^{-5}$ \\
	Restart epochs & $150, 300, 450$\\
	Learning rate schedule &
	$\eta(\text{epoch}) = \max(10^{-3},10^{-\lfloor\text{epoch} /
150\rfloor})$\\
     Momentum schedule &
	$c(\text{epoch}) = \min(1,10^{-2 + \lfloor\text{epoch} /
150\rfloor})$\\
	\hline
	Data & CIFAR100 \\
	Total epochs & 1000 \\
	L1 weight & $3 \times 10^{-5}$ \\
     Restart epochs & $150, 300, 450$\\
	Learning rate schedule &
	$\eta(\text{epoch}) = \max(10^{-3},10^{-\lfloor\text{epoch} /
150\rfloor})$\\
     Momentum schedule &
	$c(\text{epoch}) = \min(1,10^{-2 + \lfloor\text{epoch} /
150\rfloor})$\\
	\hline
	\multicolumn{2}{|c|}{\rigl} \\
	\hline
	Data & CIFAR10 \\
	Total epochs & 1000 \\
	Sparse initialization & erdos-renyi-kernel \\
     Density & 0.02 \\
     Prune rate & 0.3 \\
     Decay schedule & cosine \\ 
     Apply when & step end \\
     Interval & 100 \\
     End when & 65918 \\
     Learning rate & $0.1$ \\
     Momentum & $0.9$ \\
     Weight decay& $10^{-4}$\\
     Label smoothing& $0.1$\\
     Decay frequency& $20000$\\
     Warmup steps& $1760$\\
     Decay factor& $0.2$\\
	 \hline
	 Data & CIFAR100 \\
	 Total epochs & 1000 \\
	 Sparse initialization & erdos-renyi-kernel \\
      Density & 0.02 \\
      Prune rate & 0.3 \\
      Decay schedule & cosine \\
      Apply when & step end \\
      Interval & 100 \\
      End when & 65918 \\
      Learning rate & $0.1$\\
      Momentum& $0.9$\\
      Weight decay& $ 10^{-4}$\\
      Label smoothing& $0.1$\\
      Decay frequency& $20000$\\
      Warmup steps& $1760$\\
      Decay factor& $0.2$\\
      \hline
\end{tabular}
\end{table}

\section{More Results from the Experiments}
\begin{figure}[tb]
\begin{center}
\begin{subfigure}[b]{0.45\textwidth}
	\centering
	\includegraphics[width=\textwidth]{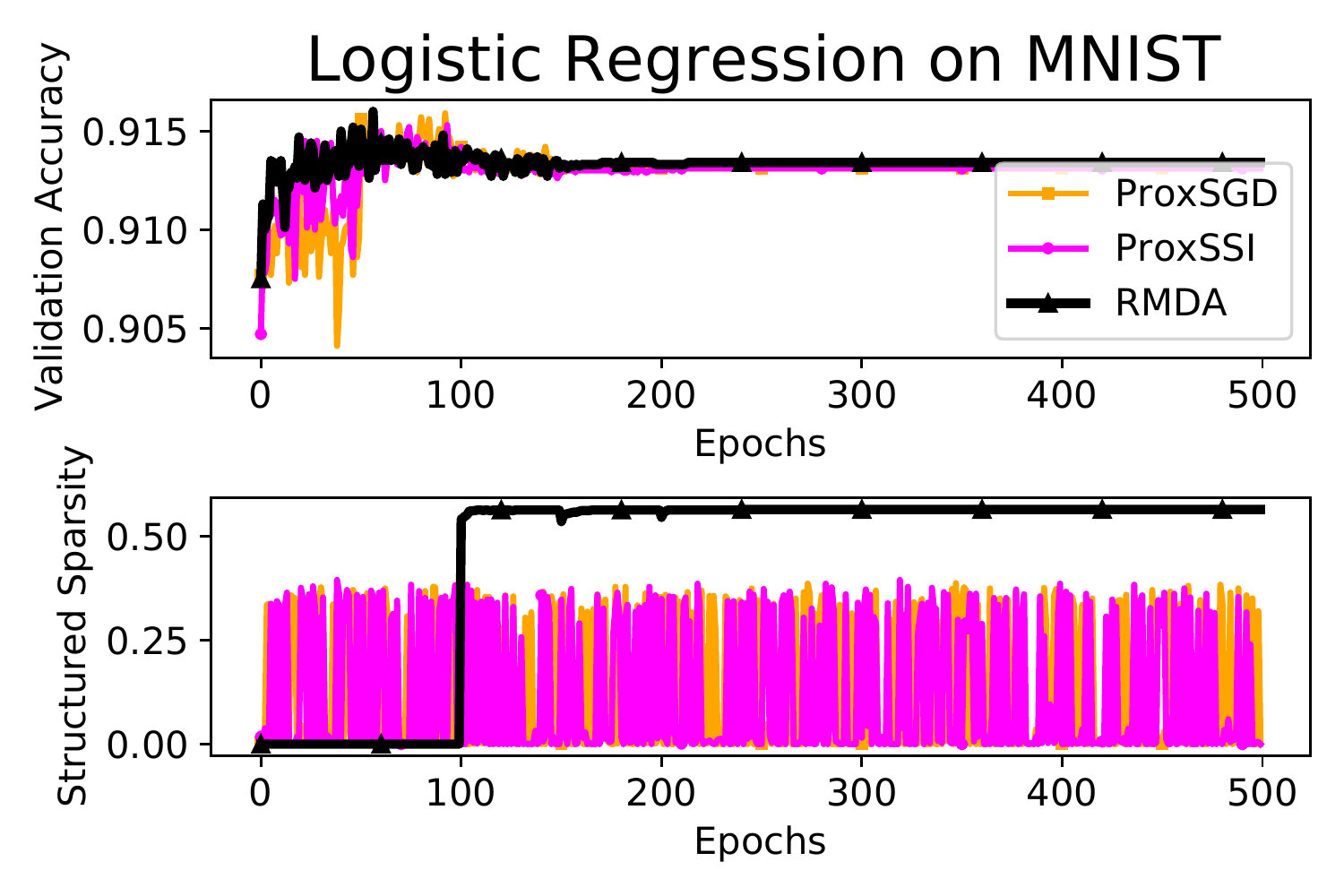}
	\caption{Logistic regression on MNIST}
\end{subfigure}
\begin{subfigure}[b]{0.45\textwidth}
	\centering
	\includegraphics[width=\textwidth]{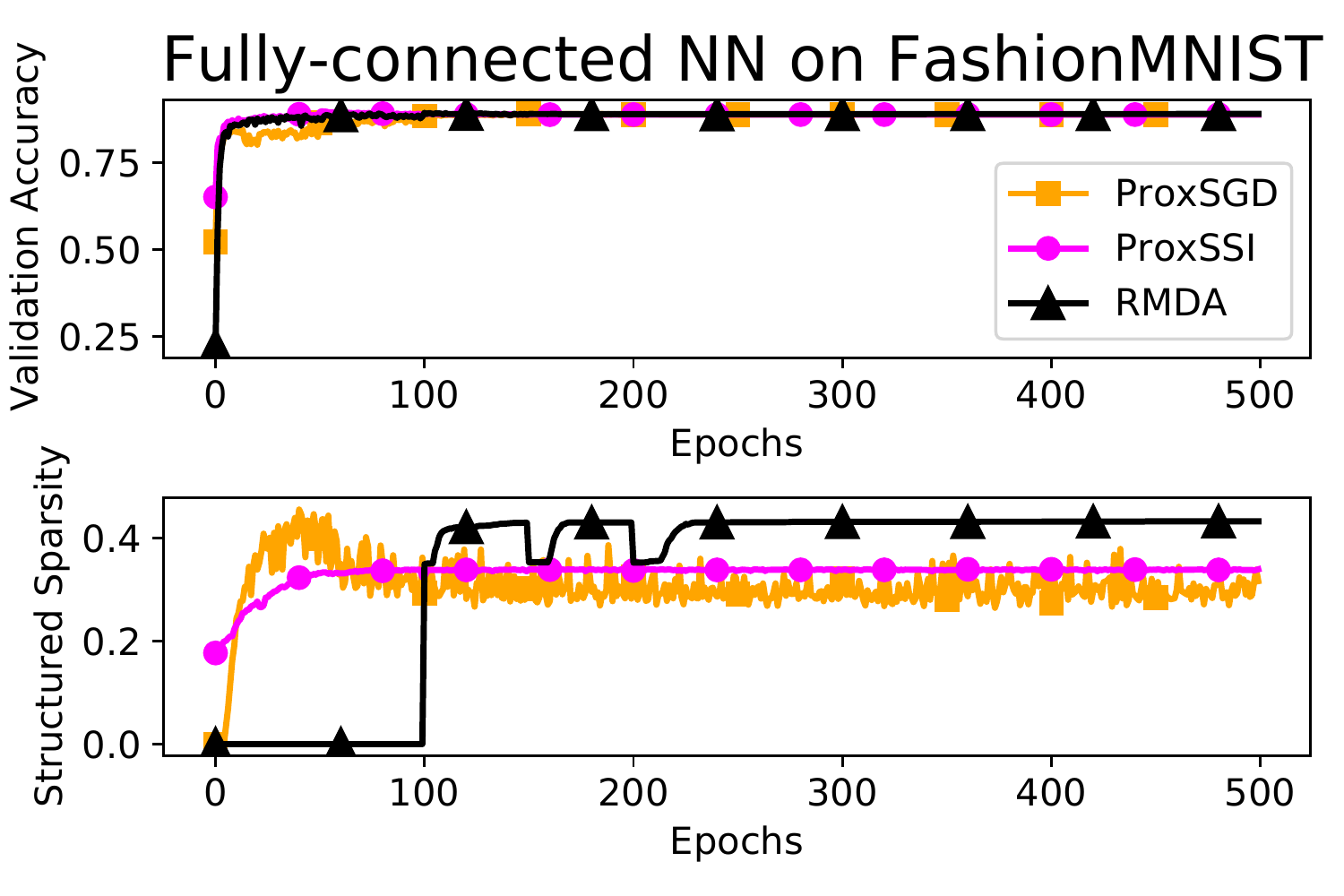}
	\caption{Fully-connected NN on FashionMNIST}
\end{subfigure}
\begin{subfigure}[b]{0.45\textwidth}
	\centering
	\includegraphics[width=\textwidth]{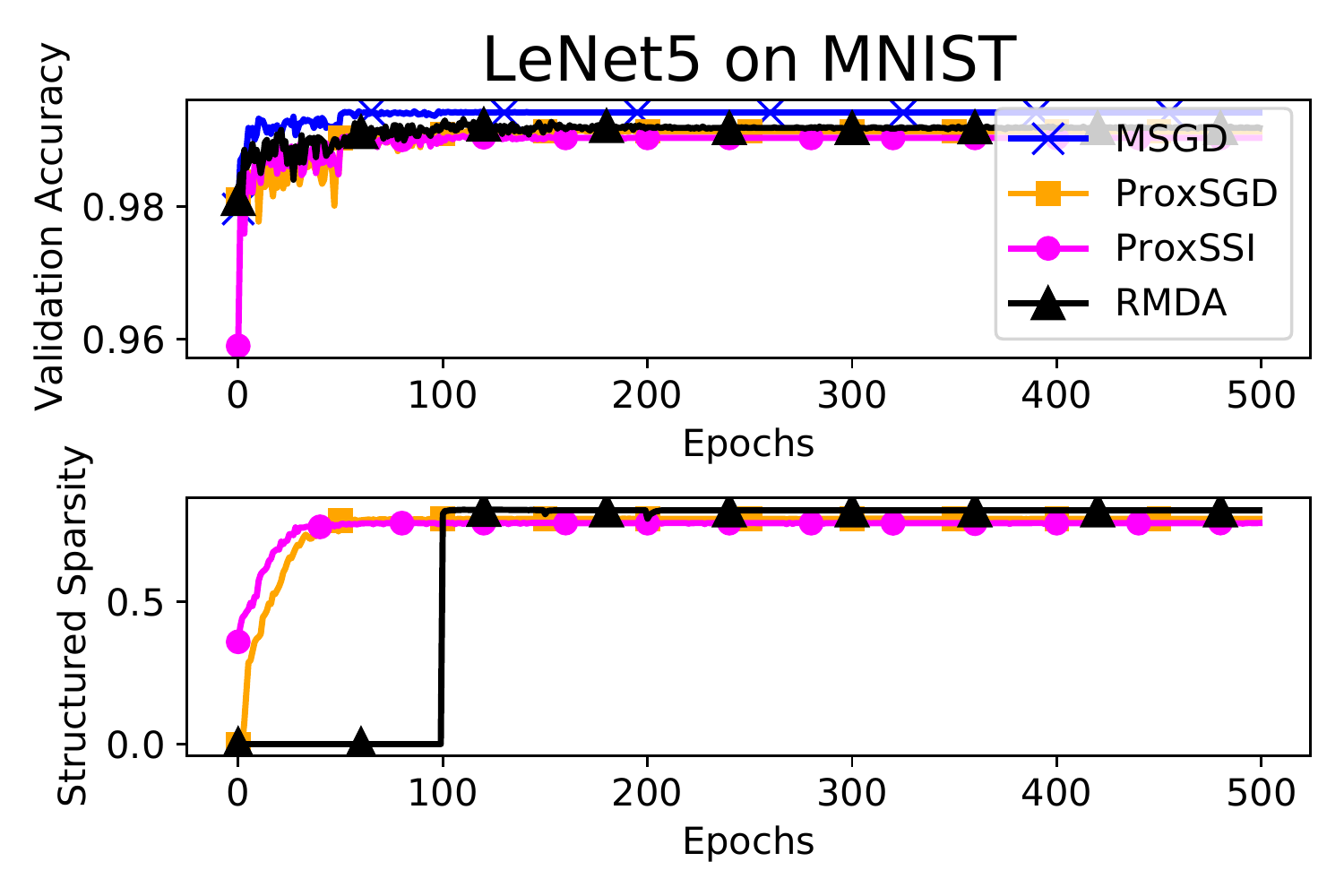}
	\caption{LeNet5 on MNIST}
\end{subfigure}
\begin{subfigure}[b]{0.45\textwidth}
	\centering
	\includegraphics[width=\textwidth]{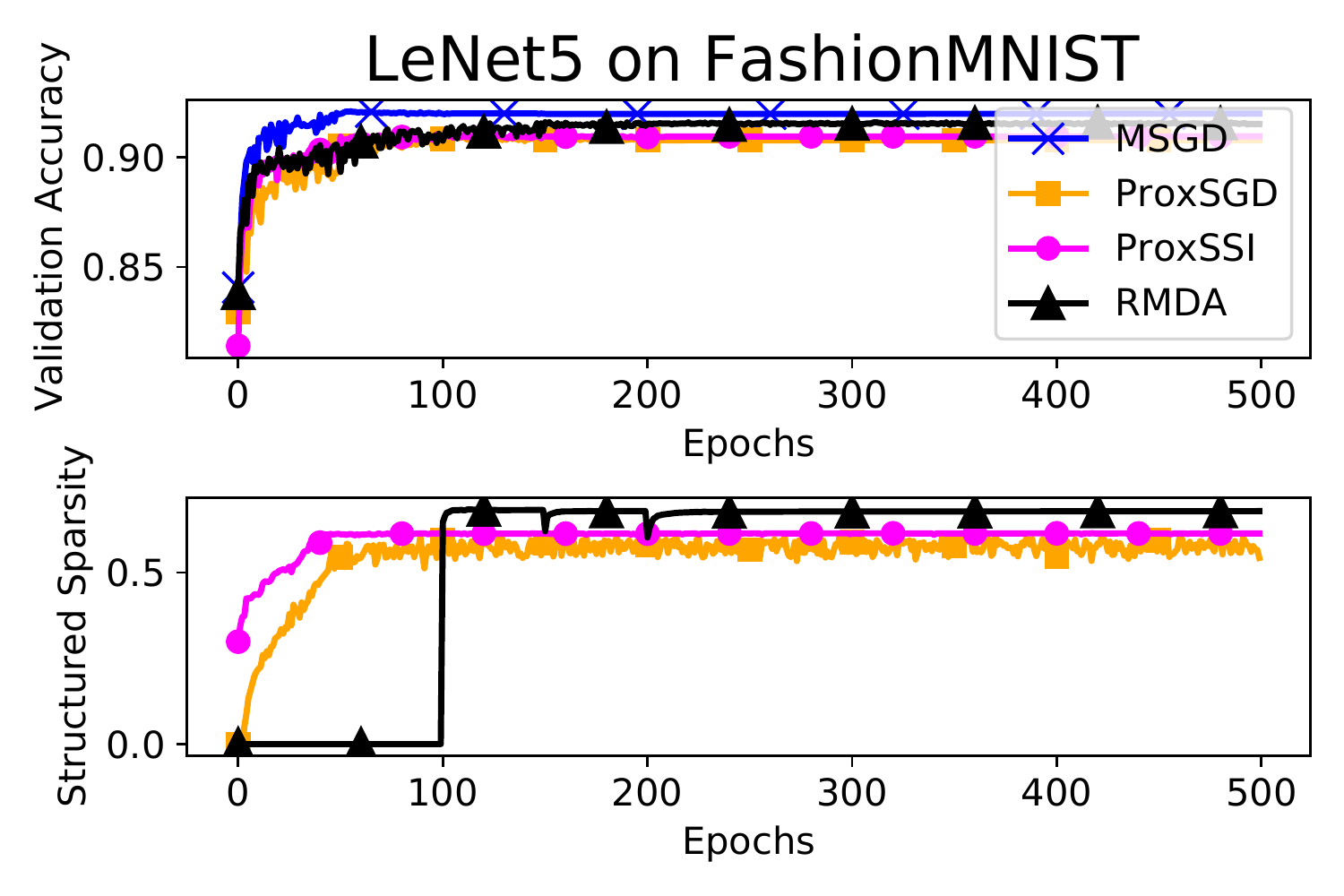}
	\caption{LeNet5 on FashionMNIST}
\end{subfigure}
\begin{subfigure}[b]{0.45\textwidth}
	\centering
	\includegraphics[width=\textwidth]{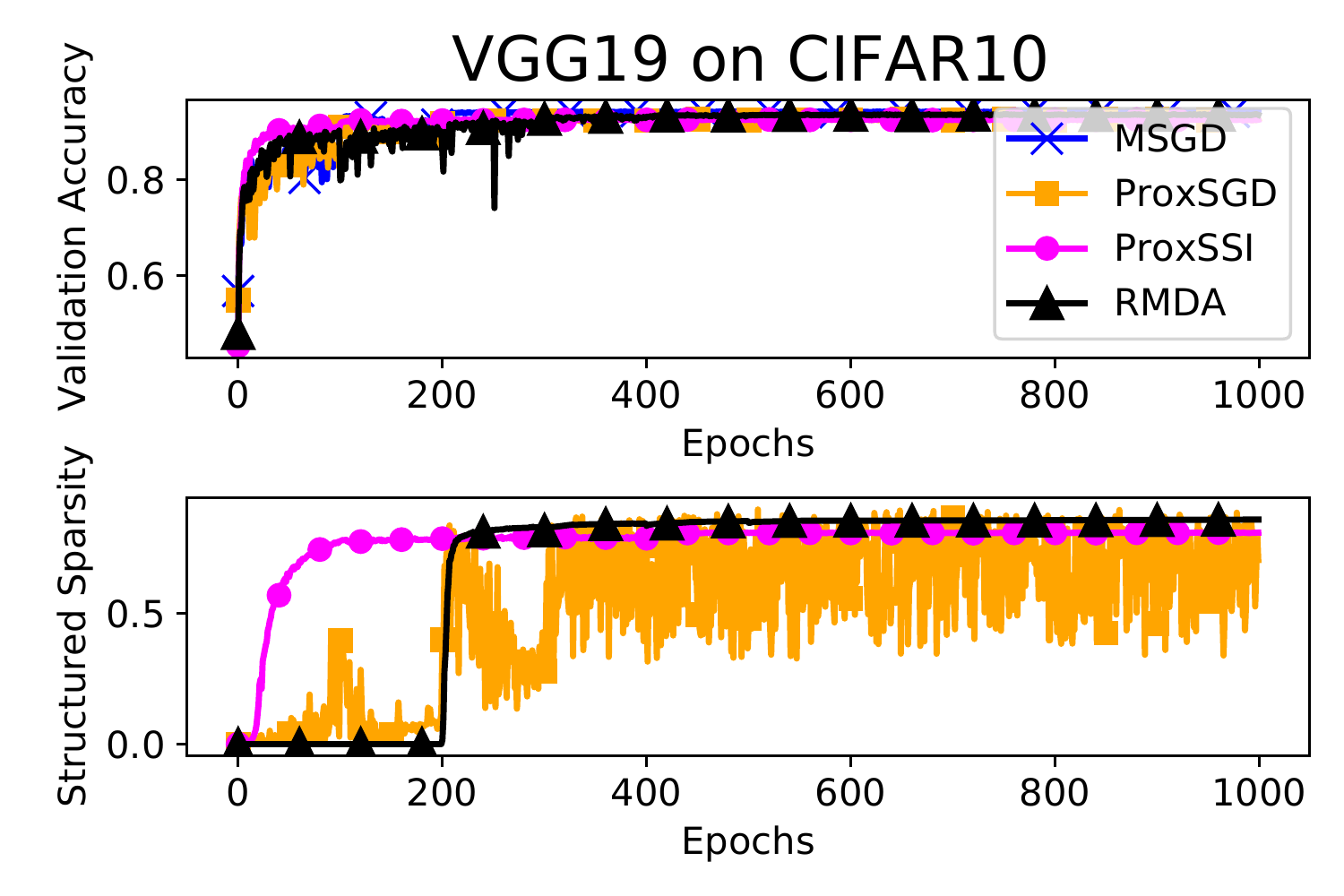}
	\caption{VGG19 on CIFAR10}
\end{subfigure}
\begin{subfigure}[b]{0.45\textwidth}
	\centering
	\includegraphics[width=\textwidth]{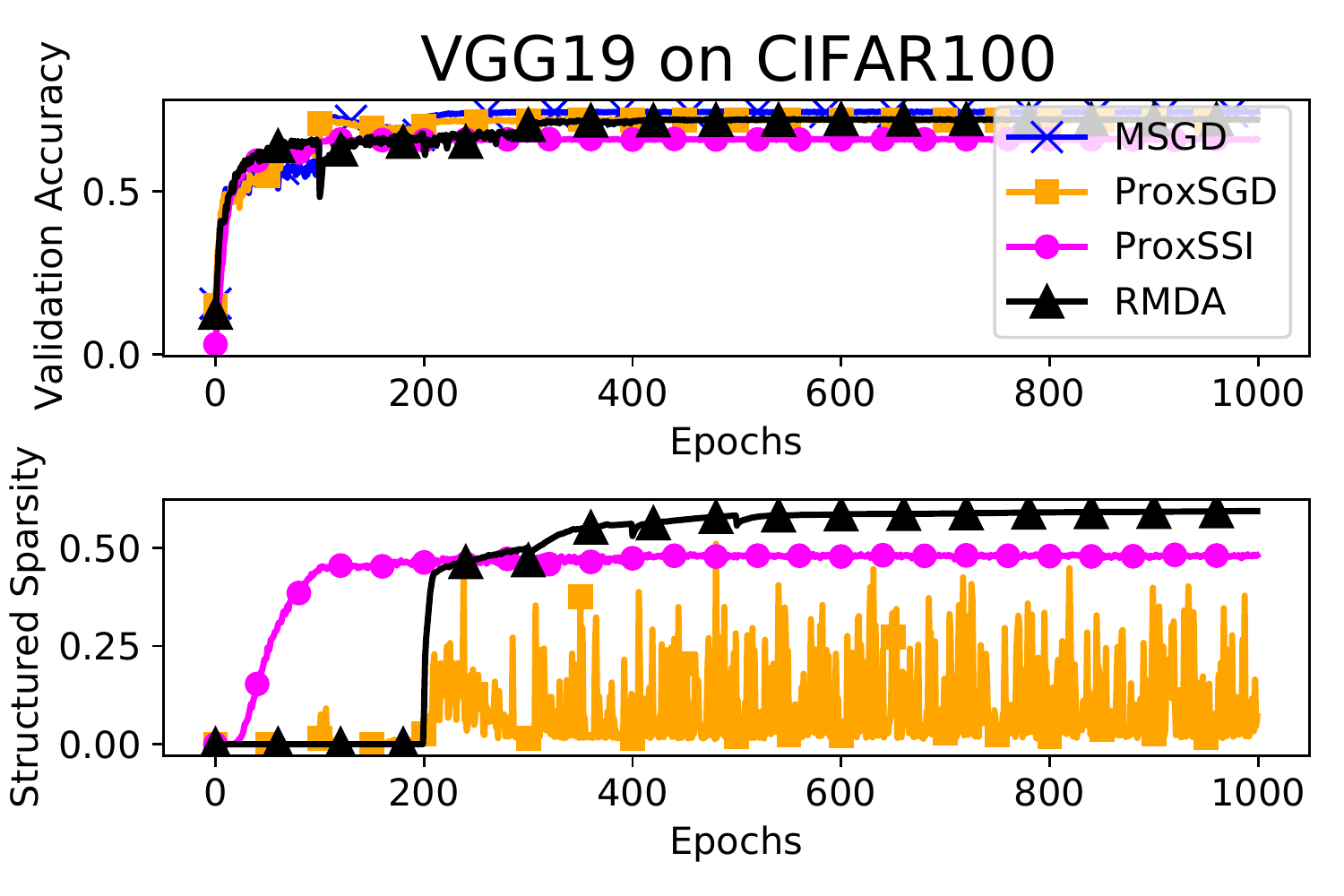}
	\caption{VGG19 on CIFAR100}
\end{subfigure}
\begin{subfigure}[b]{0.45\textwidth}
	\centering
	\includegraphics[width=\textwidth]{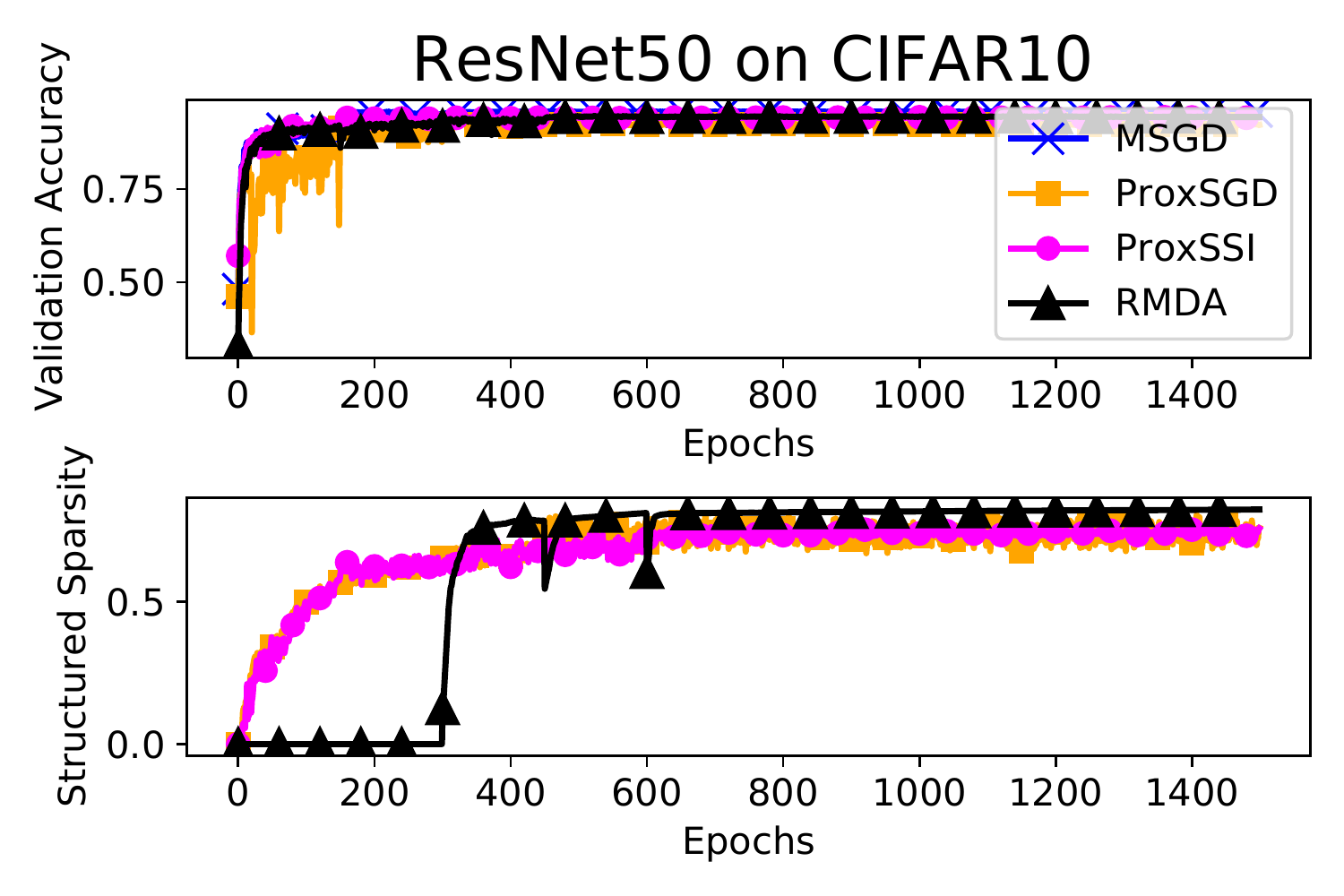}
	\caption{ResNet50 on CIFAR10}
\end{subfigure}
\begin{subfigure}[b]{0.45\textwidth}
	\centering
	\includegraphics[width=\textwidth]{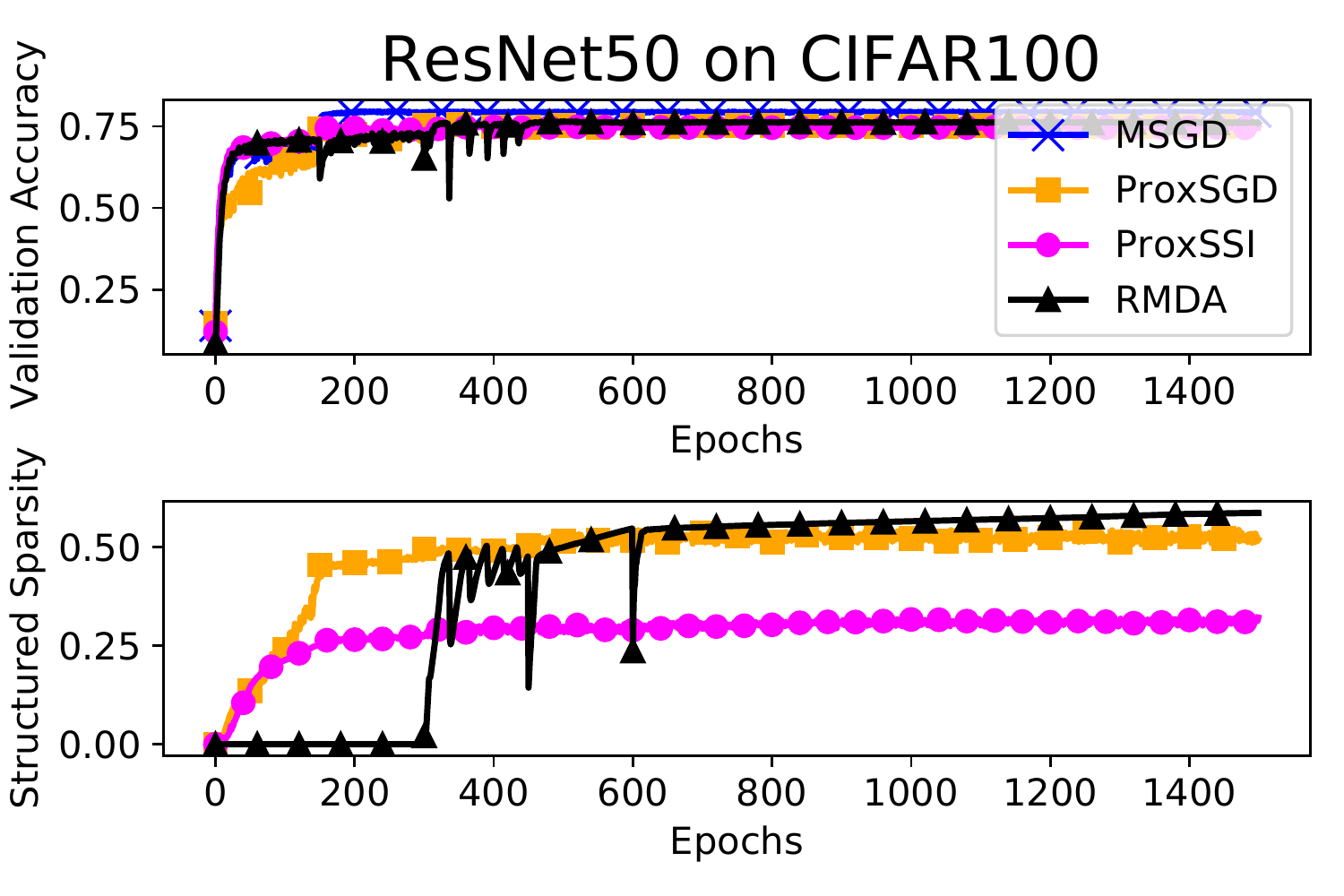}
	\caption{ResNet50 on CIFAR100}
\end{subfigure}
\end{center}
\caption{Group Sparsity and validation accuracy v.s epochs of
different algorithms on various models with the group-LASSO regularization of a single run.}
 \label{fig:epoch1}
\end{figure}

\begin{figure}[tb]
\begin{center}
\begin{subfigure}[b]{0.45\textwidth}
	\centering
	\includegraphics[width=\textwidth]{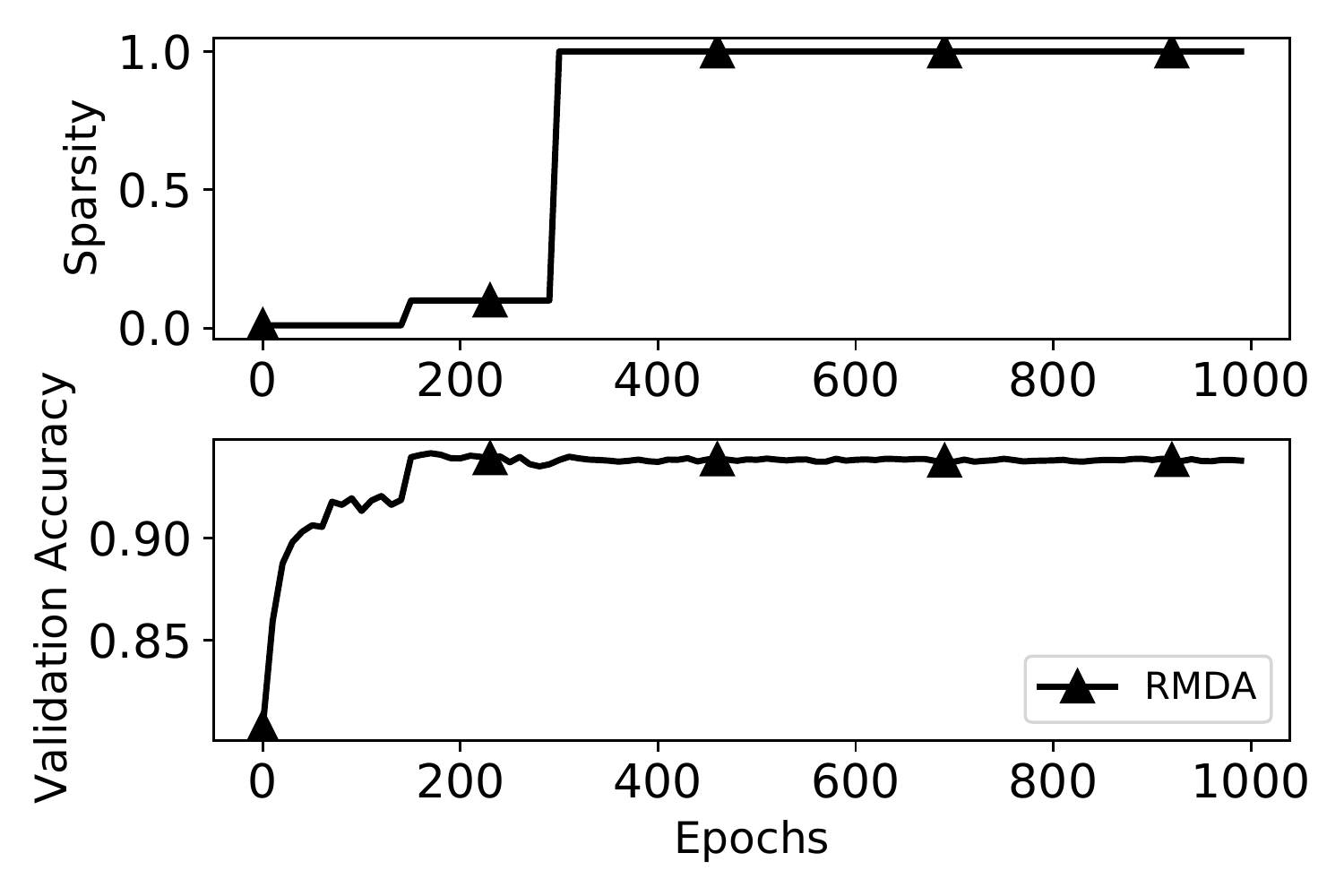}
	\caption{ResNet50 on CIFAR10}
\end{subfigure}
\begin{subfigure}[b]{0.45\textwidth}
	\centering
	\includegraphics[width=\textwidth]{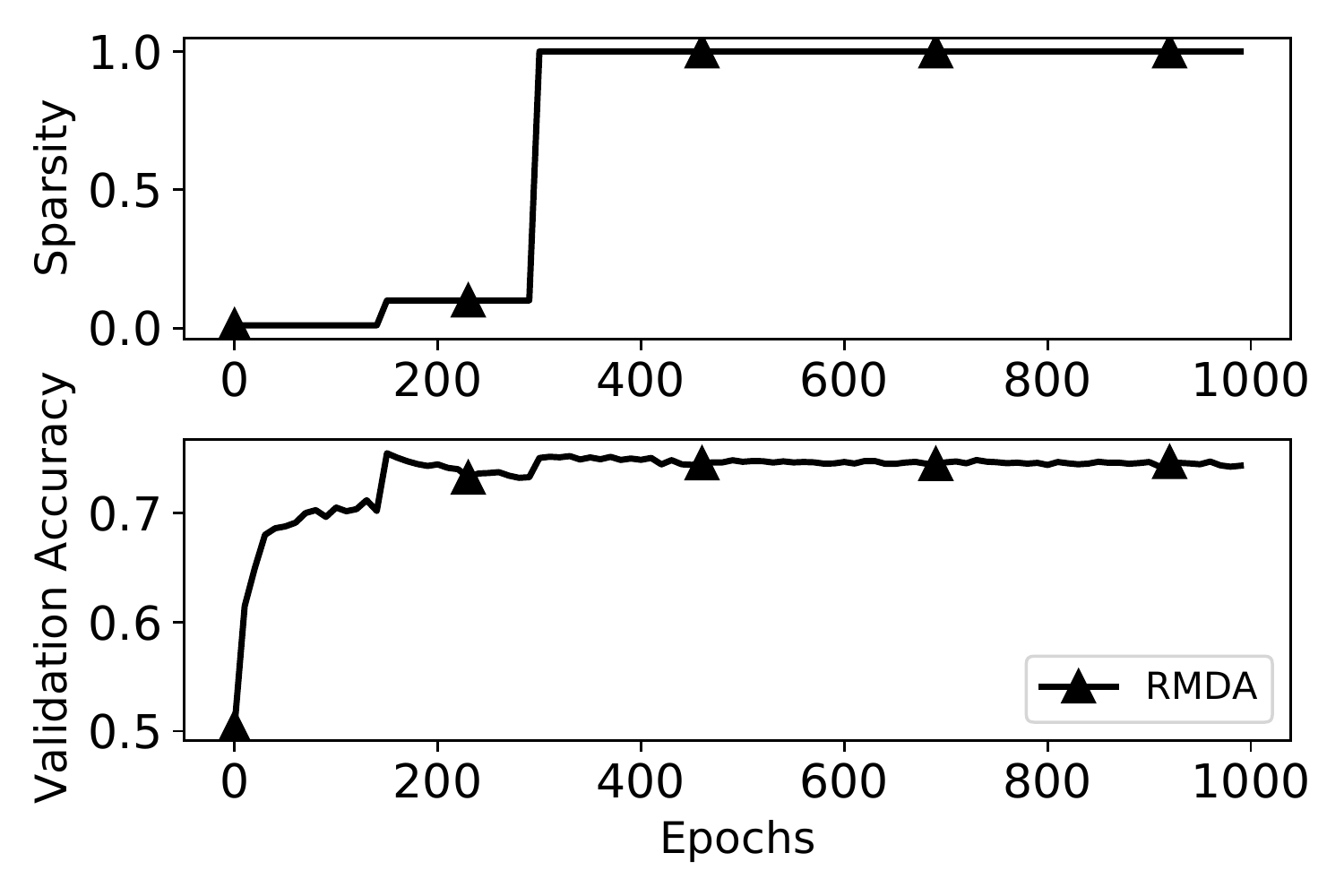}
	\caption{ResNet50 on CIFAR100}
\end{subfigure}
\end{center}
\caption{Unstructured Sparsity and validation accuracy v.s epochs of
RMDA on ResNet50 of a single run.}
 \label{fig:epoch2}
\end{figure}

In this section, we provide more details of the results of the
experiments we conducted in the main text.
In particular, in \cref{fig:epoch1}, we present the change of
validation accuracies and group sparsity levels with epochs for the
group sparsity tasks in \cref{sec:NN}.
We then present in \cref{fig:epoch2} validation accuracies and
unstructured sparsity level versus epochs for the task in
\cref{sec:cp}.
We note that although it takes more epochs for \rmda to fully
stabilize in terms of manifold identification, the sparsity level
usually only changes in a very limited range once (sometimes even
before) the validation accuracy becomes steady, meaning that we do not
need to run the algorithm for an unreasonably long time to obtain
satisfactory results.

\section{Other Regularizers for Possibly Better Group Sparsity and Generalization}
\label{sec:mcp}
A downside of \cref{eq:grouplasso} is that it pushes all groups toward
zeros and thus introduces bias in the final model.
For its remedy, minimax concave penalty \citep[MCP,][]{zhang2010nearly}
is then proposed to penalize only the groups whose norm is smaller
than a user-specified threshold.
More precisely, given hyperparameters $\lambda \geq 0, \omega \geq 1$, the
one-dimensional MCP is defined by
\begin{equation*}
	\MCP(w;\lambda,\omega)  \coloneqq
	\begin{cases}
	\lambda |w| - \frac{w^2}{2 \omega} & \text{if} |w| < \omega
	\lambda,\\
	\frac{\omega {\lambda}^{2}}{2} & \text{if}  |w| \geq \omega \lambda.
	\end{cases}
\end{equation*}
One can then apply the above formulation to the norm of a vector to
achieve the effect of inducing group-sparsity.
In our case, given an index set $\gI_g$ that represents a group, the
MCP for this group is then computed as
\citep{breheny2009penalized}
\begin{equation*}
\MCP\left(W_{\gI_g}; \lambda_g, \omega_g\right) \coloneqq
\begin{cases}
\lambda_g \norm{W_{\gI_g}}^2 - \frac{\norm{W_{\gI_g}}^2}{2 \omega_g} &
\text{if} \  \norm{W_{\gI_g}} < \omega_g \lambda_g, \\
\frac{\omega_g {\lambda_g}^{2} }{2} & \text{if}
\norm{W_{\gI_g}} \geq \omega_g \lambda_g.
\end{cases}
\end{equation*}
We then consider
\begin{equation}
	\label{eq:mcp}
	\psi(W) = \sum_{g=1}^{|\gG|} MCP\left(W_{\gI_g}; \lambda_g,\omega_g\right).
\end{equation}

\begin{figure}
	\begin{center}
		\begin{tabular}{cc}
			\includegraphics[width=0.48\linewidth]{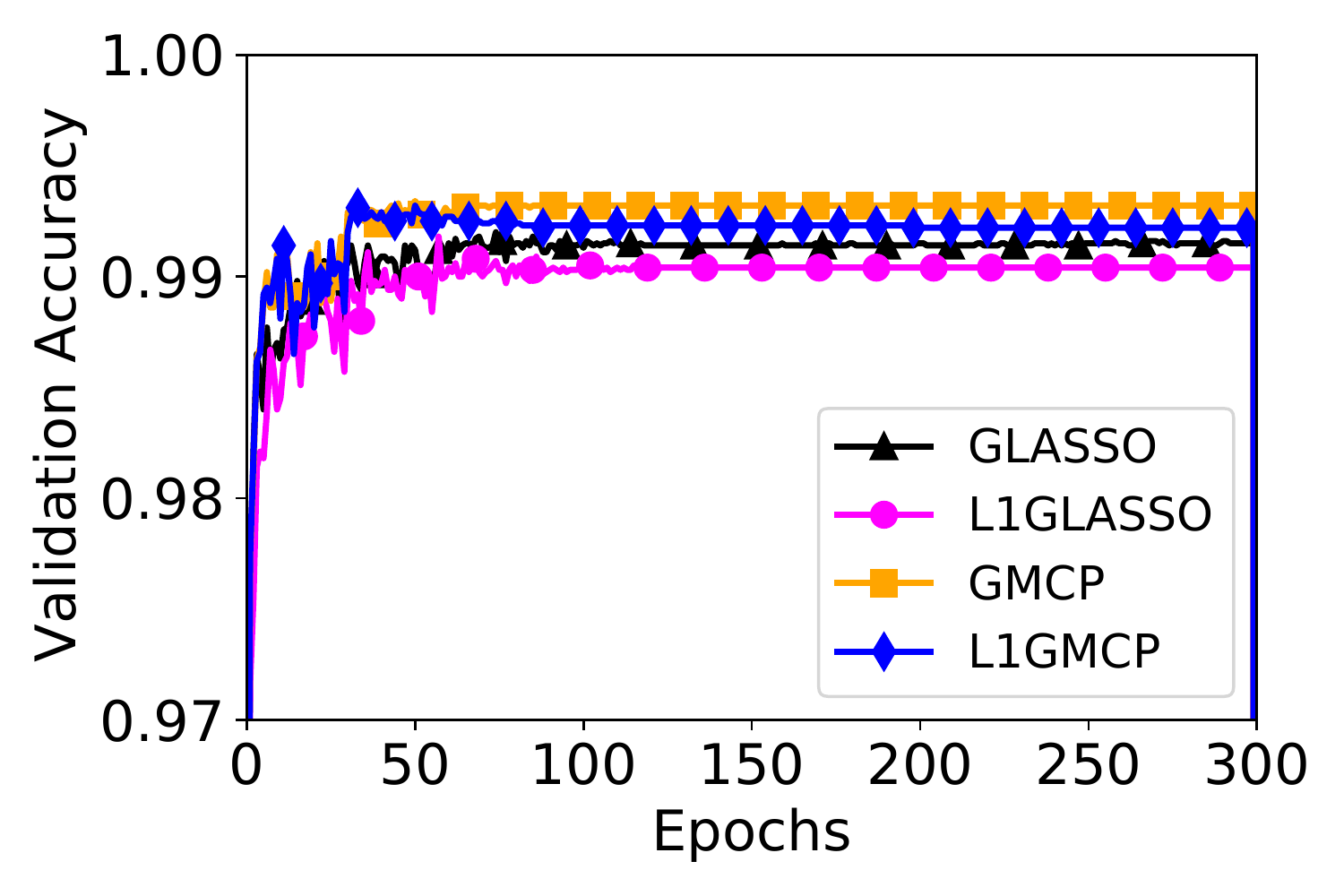}&
			\includegraphics[width=0.48\linewidth]{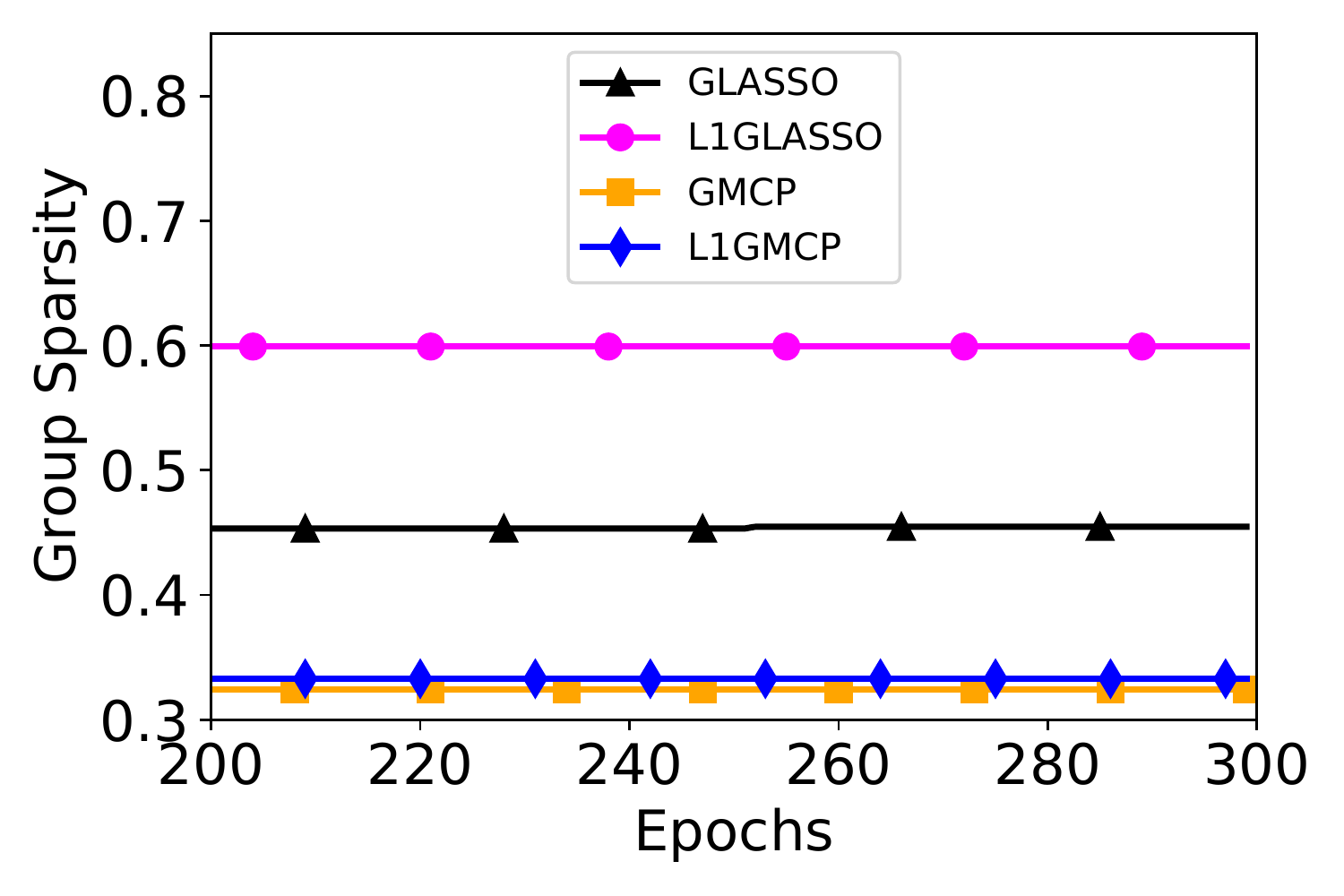}
		\end{tabular}
	\end{center}
     \caption{Comparison between group LASSO, L1+group LASSO, Group MCP, L1+Group MCP}
     \label{fig:LeNet5_reg}
\end{figure}

It is shown in \cite{deleu2021structured} that group MCP
regularization may simultaneously provide higher group sparsity and
better validation accuracy than the group LASSO norm in vision and
language tasks.
Another possibility to enhance sparsity is to add another
$\ell_1$-norm or entry-wise MCP regularization to the group-level
regularizer.
The major drawback of these approaches is the requirement of
additional hyperparameters, and we prefer simpler approaches over
those with more hyperparameters, as hyperparameter
tuning in the latter can be troublesome for users with limited
computational resources,
and using a simpler setting can also help us to focus on the
comparison of the algorithms themselves.
The experiment in this subsection is therefore only for illustrating
that these more complicated regularizers can be combined with \rmda if
the user wishes, and such regularizers might lead to better results.
Therefore, we train a version of LeNet5, which is slightly simpler than the one
we used in previous experiments, on the MNIST dataset with
such regularizers using \rmda and display the respective performance of
various regularization schemes in \cref{fig:LeNet5_reg}.
For the weights $w_g$ of each group in \cref{eq:grouplasso}, in this
experiment we consider the following setting.
Let $L_i$ be the collection of all index sets that belong to the $i$-th
layer in the network,
and denote
\[
	N_{L_i} \coloneqq \sum_{\gI_j \in L_i} |\gI_j|
\]
the number of parameters in this layer,
for all $i$, we set $w_g = \sqrt{N_{L_i}}$ for all $g$ such that
$\gI_g \in L_i$.
Given two constants $\lambda > 0$ and ,$\omega > 1$,
The values of $\lambda_g$ and $\omega_g$ in \cref{eq:mcp} are then
assigned as $\lambda_g = \lambda w_g$ and $\omega_g = \omega w_g$.

In this figure, group LASSO is abbreviated as GLASSO;   $\ell_1$-norm
plus a group LASSO norm, L1GLASSO;
group MCP, GMCP; element-wise MCP plus group MCP, L1GMCP.
Our results exemplify that different regularization schemes might have
different benefits on one of the criteria with proper hyperparameter
tuning.
The detailed numbers are reported in \cref{tbl:diffreg_results} and
the experiment settings can be found in \cref{tbl:reg,tbl:LeNet5_small}.

\begin{table}[tbh]
\caption{Results of training LeNet5 on MNIST using \rmda with
different regularizers. We report mean and standard deviation of three
independent runs.}
\label{tbl:diffreg_results}
\sisetup{detect-weight,mode=text}
\renewrobustcmd{\bfseries}{\fontseries{b}\selectfont}
\renewrobustcmd{\boldmath}{}
\newrobustcmd{\B}{\bfseries}
\begin{center}
	\begin{tabular}{lrr}
		\toprule
		Regularizers & Validation accuracy & Group sparsity\\
		\midrule
		GLASSO & 99.11 $\pm$ 0.06\% & 45.33 $\pm$ 0.99\% \\
		L1GLASSO & 99.02 $\pm$ 0.01\% & 58.92 $\pm$ 1.30\% \\
		GMCP & 99.25 $\pm$ 0.08\% & 32.81 $\pm$ 0.96\% \\
		L1GMCP & 99.21 $\pm$ 0.03\% & 32.91 $\pm$ 0.35\% \\
		\hline
	\end{tabular}
\end{center}
\end{table}

\begin{table}[tbh]
\centering
\caption{Details of the modified simpler LeNet5 for the experiment in
	\cref{sec:mcp}.
\url{https://github.com/zihsyuan1214/rmda/blob/master/Experiments/Models/lenet5_small.py}.}
\label{tbl:LeNet5_small}
\begin{tabular}{|l|l|}
	\hline  
	Parameter & Value \\\hline
	Number of layers & 5 \\
	Number of convolutional layers & 3 \\
	Number of fully-connected layers & 2 \\
	Size of convolutional kernels & $5 \times 5$ \\
	Number of output filters $1,2$ & $6, 16$ \\
	Number of output neurons $3, 4, 5$ & $120, 84, 10$ \\
	Kernel size, stride, padding of maxing pooling & $2 \times 2$, none, invalid\\
	Operations after convolutional layers & max pooling \\
	Activation function for convolution/output layer & relu/softmax \\
	\hline
\end{tabular}
\end{table}

\begin{table}[tbh]
\centering
\caption{Details of the experimental settings for comparing different
	regularizers in \cref{sec:mcp}}
\label{tbl:reg}
\begin{tabular}{|l|l|}
	\hline  
	Parameter & Value \\\hline
	Data set & MNIST \\
	Model & LeNet5 (\cref{tbl:LeNet5_small}) \\
	Loss function & Cross entropy \\
	Algorithms & \rmda \\
	Total epochs & 300 \\
	Restart epochs & $30, 60, 90, 120$\\
	Learning rate schedule &
	$\eta(\text{epoch}) = \max(10^{-5},10^{-1 - \lfloor\text{epoch} /
30\rfloor})$\\
     Momentum schedule & $c(\text{epoch}) = \min(1,10^{-2 +
		 \lfloor\text{epoch} / 30\rfloor})$\\
	\hline
	\multicolumn{2}{|c|}{GLASSO} \\
	\hline
	Group LASSO weight& $10^{-5}$ \\
	\hline
	\multicolumn{2}{|c|}{L1GLASSO } \\
	\hline
	L1 weight & $10^{-4}$ \\
	Group LASSO weight & $10^{-5}$ \\
	\hline
	\multicolumn{2}{|c|}{GMCP} \\
	\hline
	Group MCP weight & $10^{-5}$ \\
	$\gamma$ & $64$ \\
	\hline	
	\multicolumn{2}{|c|}{L1GMCP} \\
	\hline
	L1 weight & $10^{-4}$ \\
	Group MCP weight & $10^{-5}$ \\
	$\gamma$ & $64$ \\
     \hline
\end{tabular}
\end{table}

\end{document}

%% file: math_commands.tex
\usepackage{amsmath,amsfonts,bm}

\def\eqref#1{equation~\ref{#1}}

\def\1{\bm{1}}

\DeclareMathAlphabet{\mathsfit}{\encodingdefault}{\sfdefault}{m}{sl}
\SetMathAlphabet{\mathsfit}{bold}{\encodingdefault}{\sfdefault}{bx}{n}

\def\gA{{\mathcal{A}}}
\def\gB{{\mathcal{B}}}

\def\gD{{\mathcal{D}}}
\def\gE{{\mathcal{E}}}
\def\gF{{\mathcal{F}}}
\def\gG{{\mathcal{G}}}

\def\gI{{\mathcal{I}}}

\def\gL{{\mathcal{L}}}
\def\gM{{\mathcal{M}}}

\def\gR{{\mathcal{R}}}

\def\gZ{{\mathcal{Z}}}

\newcommand{\E}{\mathbb{E}}

\newcommand{\R}{\mathbb{R}}

\DeclareMathOperator*{\argmin}{arg\,min}